\documentclass[sigconf]{acmart}

\AtBeginDocument{%
  }

\copyrightyear{2025}
\acmYear{2025}
\setcopyright{cc}
\setcctype{by}
\acmConference[SC '25]{The International Conference for High Performance Computing, Networking, Storage and Analysis}{November 16--21, 2025}{St Louis, MO, USA}
\acmBooktitle{The International Conference for High Performance Computing, Networking, Storage and Analysis (SC '25), November 16--21, 2025, St Louis, MO, USA}
\acmDOI{10.1145/3712285.3759890}
\acmISBN{979-8-4007-1466-5/2025/11}

\usepackage{amsmath,amsfonts,amsmath}
\usepackage{graphicx}
\usepackage{textcomp}
\usepackage{xcolor}
\usepackage{listings}
\usepackage{xspace}
\usepackage{algorithm}
\usepackage{algpseudocode}
\usepackage{bm}
\numberwithin{equation}{section}

\usepackage{soul}
\usepackage{annotate-equations}
\usepackage{ctable}

\definecolor{dkgreen}{RGB}{0,64,0}
\definecolor{ltgray}{RGB}{245,245,245}
\definecolor{mauve}{RGB}{139,0,139}

\definecolor{color_1}{HTML}{009E73} 
\definecolor{color_2}{HTML}{D55E00} 
\definecolor{color_3}{HTML}{0072B2} 
\definecolor{color_4}{HTML}{000000} 
\definecolor{color_5}{HTML}{E69F00} 
\definecolor{color_6}{HTML}{CC79A7} 

\colorlet{soul_1}{color_1!20}
\colorlet{soul_2}{color_2!20}
\colorlet{soul_3}{color_3!20}
\colorlet{soul_4}{color_4!20}
\colorlet{soul_5}{color_5!20}
\colorlet{soul_6}{color_6!20}

\newcommand{\highlight}[2]{\sethlcolor{#1}\hl{#2}}

\newcommand{\tweakedsim}{\raise.17ex\hbox{$\scriptstyle\mathtt{\sim}$}}

\newcommand{\plexus}{Plexus\xspace}

\begin{document}

\title{Plexus: Taming Billion-edge Graphs with 3D Parallel Full-graph GNN Training}

\author{Aditya K.~Ranjan}
\affiliation{
  \institution{Department of Computer Science\\University of Maryland}
  \city{College Park}
  \state{Maryland}
  \country{USA}
}
\email{aranjan2@umd.edu}
\orcid{0009-0000-5390-7800}

\author{Siddharth Singh}
\affiliation{
  \institution{Department of Computer Science\\University of Maryland}
  \city{College Park}
  \state{Maryland}
  \country{USA}
}
\email{ssingh37@umd.edu}
\orcid{0000-0002-2756-4290}

\author{Cunyang Wei}
\affiliation{
  \institution{Department of Computer Science\\University of Maryland}
  \city{College Park}
  \state{Maryland}
  \country{USA}
}
\email{cunyang@umd.edu}
\orcid{0009-0001-8910-4951}

\author{Abhinav Bhatele}
\affiliation{
  \institution{Department of Computer Science\\University of Maryland}
  \city{College Park}
  \state{Maryland}
  \country{USA}
}
\email{bhatele@cs.umd.edu}
\orcid{0000-0003-3069-3701}

\renewcommand{\shortauthors}{Ranjan et al.}

\begin{abstract}
Graph neural networks (GNNs) leverage the connectivity and structure of
real-world graphs to learn intricate properties and relationships between
nodes. Many real-world graphs exceed the memory capacity of a GPU due to their
sheer size, and training GNNs on such graphs requires techniques such as
mini-batch sampling to scale. The alternative approach of distributed
full-graph training suffers from high communication overheads and load
imbalance due to the irregular structure of graphs. We propose a
three-dimensional (3D) parallel approach for full-graph training that tackles
these issues and scales to billion-edge graphs. In addition, we introduce
optimizations such as a double permutation scheme for load balancing, and a
performance model to predict the optimal 3D configuration of our parallel
implementation -- \plexus. We evaluate \plexus on six different graph datasets
and show scaling results on up to 2048 GPUs of Perlmutter, and 1024 GPUs of
Frontier. \plexus achieves unprecedented speedups of 2.3$-$12.5$\times$ over
prior state of the art, and a reduction in time-to-solution by
5.2$-$8.7$\times$ on Perlmutter and 7.0$-$54.2$\times$ on Frontier.

\end{abstract}



\maketitle

\section{Motivation}
\label{sec:intro}
Graphs are used to represent irregular structures and connections that are
ubiquitous in the real-world, such as molecular structures, social networks,
and financial transaction networks. In recent years, graph neural networks
(GNNs) have emerged as a powerful class of neural networks capable of
leveraging the inherent expressiveness of graphs to learn complex properties
and relationships within them. Among GNNs, the Graph Convolutional Network
(GCN)~\cite{KipfW16} is the most popular and widely adopted, and serves as the
foundation for numerous extensions, including the Graph Attention Network
(GAT)~\cite{GAT} and the Graph Isomorphism Network (GIN)~\cite{GIN}. Unlike
traditional convolutional neural networks~\cite{lenet}, which operate on
fixed-size neighborhoods, GCNs exploit the irregular structure and connectivity
of graphs.

Real-world graphs are often extremely large, and datasets representing them
frequently exceed the memory capacity of a single GPU. Kipf et
al.~\cite{KipfW16} recognize this limitation of their seminal work and suggest
mini-batch training for scaling to larger graphs, where a small subset of nodes
is used in each iteration to update the model. Since efficient and scalable
full-graph based approaches are missing, most modern frameworks such as PyTorch
Geometric~\cite{pytorchgeometric} and DGL~\cite{dgl} use mini-batch training as
their default.

In mini-batch training, in a single GCN layer, nodes in the mini-batch first
collect information from their immediate neighbors. By aggregating feature
embeddings from a node's neighborhood and applying a feed-forward
transformation, GCNs can address tasks such as node-level, link-level, and
graph-level predictions. For a model with $K$ such GCN layers, a node
aggregates features from its $K$-hop neighborhood.  However, even for small
values of $K$, this can quickly result in a phenomenon known as neighborhood
explosion, accessing large portions of the graph and undermining the efficiency
of mini-batch training~\cite{chen2018stochastictraininggraphconvolutional}. To
mitigate this issue, sampling algorithms such as GraphSAGE~\cite{graphsage} and
FastGCN~\cite{fastgcn} are typically applied alongside mini-batch training to
reduce the number of neighbors considered, thereby lowering memory consumption.

While sampling is widely used, it comes with inherent limitations. Most
notably, sampling introduces approximations that can lead to degradation in
accuracy~\cite{MLSYS2020ROC}. Further, CPU-GPU data transfers in sampling often
dominate training time and add unnecessary
complexity~\cite{yuan2024comprehensiveevaluationgnntraining}.  Full-graph
training, on the other hand, can achieve competitive performance without these
trade-offs in many scenarios as shown by Jia et al.'s ROC
framework~\cite{MLSYS2020ROC}.  Full-graph training makes no approximations in
the training process and avoids the complexity of choosing an appropriate
sampling strategy with suitable hyperparameters. For these reasons, in this
work, we focus on the full-graph training paradigm, avoiding any
approximations.

Graphs are typically represented as adjacency matrices with a non-zero entry
for each edge. The non-zero entries are sparsely and unevenly distributed
across the matrix. Among the six graphs we use for evaluation in this work, the
fraction of zeros in the adjacency matrix ranges from 99.79\% to 99.99\%. The
largest of these graphs has \tweakedsim 111 million vertices and \tweakedsim
1.6 billion directed edges. These characteristics of graphs introduce several
challenges in parallelizing the training.  First, high memory requirements
necessitate distributing the graph and its features, and the associated
computation across multiple GPUs.  This incurs high communication overheads due
to the need to synchronize large intermediate activations and gradients between
GPUs.  Consequently, parallel GNN training quickly becomes communication-bound,
making it difficult to scale efficiently to a large number of GPUs. Second, the
aggregation phase involves Sparse Matrix-Matrix Multiplication (SpMM), which
dominates the computational time and suffers from poor performance on GPUs due
to irregular memory access patterns and low data reuse. Third, unevenly
distributed sparsity patterns in the adjacency matrix can lead to significant
computational load imbalance across different GPUs, which can ripple through an
epoch, and impact communication times as well.

In order to address the challenges mentioned above, we propose a
three-dimensional (3D) parallel algorithm that enables scaling to large graphs
by distributing all matrices efficiently across multiple GPUs, and parallelizes
all matrix multiplication computations involved in training. Our approach draws
inspiration from Agarwal et al.'s 3D parallel matrix multiplication
algorithm~\cite{agarwal-3d}, which has been used in several distributed deep
learning frameworks, including Colossal-AI's unified deep learning
system~\cite{colossalai2023unified}, AxoNN~\cite{singh:ipdps2022, singh:sc2024}, and Eleuther
AI's framework OSLO~\cite{oslo}. We introduce several optimizations in our
baseline implementation to improve performance further.  One optimization is a
double permutation scheme, which ensures a near-perfect even distribution of
non-zeros across distributed matrices, which helps eliminate load imbalance. We
also develop a performance model that helps users select an optimal
configuration for mapping computation to a 3D virtual GPU grid. This eliminates
the need for exhaustive testing of different 3D configurations while ensuring
robust performance outcomes.



Our key contributions are summarized as follows:
\begin{itemize}
\item We present \plexus\footnote{\url{https://github.com/hpcgroup/plexus}}, an
open-source 3D parallel framework for full-graph GNN training that scales to
massive graphs and large GPU-based supercomputers.
\item A performance model to identify the optimal configuration for arranging
GPUs within a 3D virtual grid.
\item Performance optimizations, including a double permutation scheme to
mitigate load imbalance, and blocked aggregation to reduce performance
variability.
\item Unprecedented scaling to 1024 GPUs on Frontier at OLCF and 2048 GPUs on
Perlmutter at NERSC -- the largest-scale full-graph GNN training reported to
date.
\item Significant speedups, achieving 2.3$-$12.5$\times$ faster training than
state-of-the-art frameworks, and cutting time-to-solution by 5.2$-$8.7$\times$
on Perlmutter and 7.0$-$54.2$\times$ on Frontier.
\end{itemize}

\section{Background and Related Work}
\label{sec:bg}
In this section, we provide an overview of how GNNs work, different training
paradigms for GNNs, as well as challenges associated with distributed
full-graph GNN training. We also present existing GNN frameworks and their
limitations, motivating the need for our work.

\subsection{Mathematical Formulation of a GCN layer}
\label{subsec:math}

Similar to other ML models, GCNs can have different downstream tasks depending
on the application. They can be used for predicting whether an edge exists
between two nodes, predicting a holistic property of the whole graph,
predicting classes for individual nodes, etc. In this work, we focus on the
node-level classification task. However, we note that our method can be easily
be adapted to other downstream tasks as well. The primary goal of a GNN in this
setting is not only to learn a function that maps nodes to their target outputs
but also to learn high-quality, low-dimensional node embeddings that place
similar nodes close together in the embedding space. In this section, we will
show how this task is formulated using a GCN.

\begin{figure*}[t]
  \centering
  \includegraphics[height=2in]{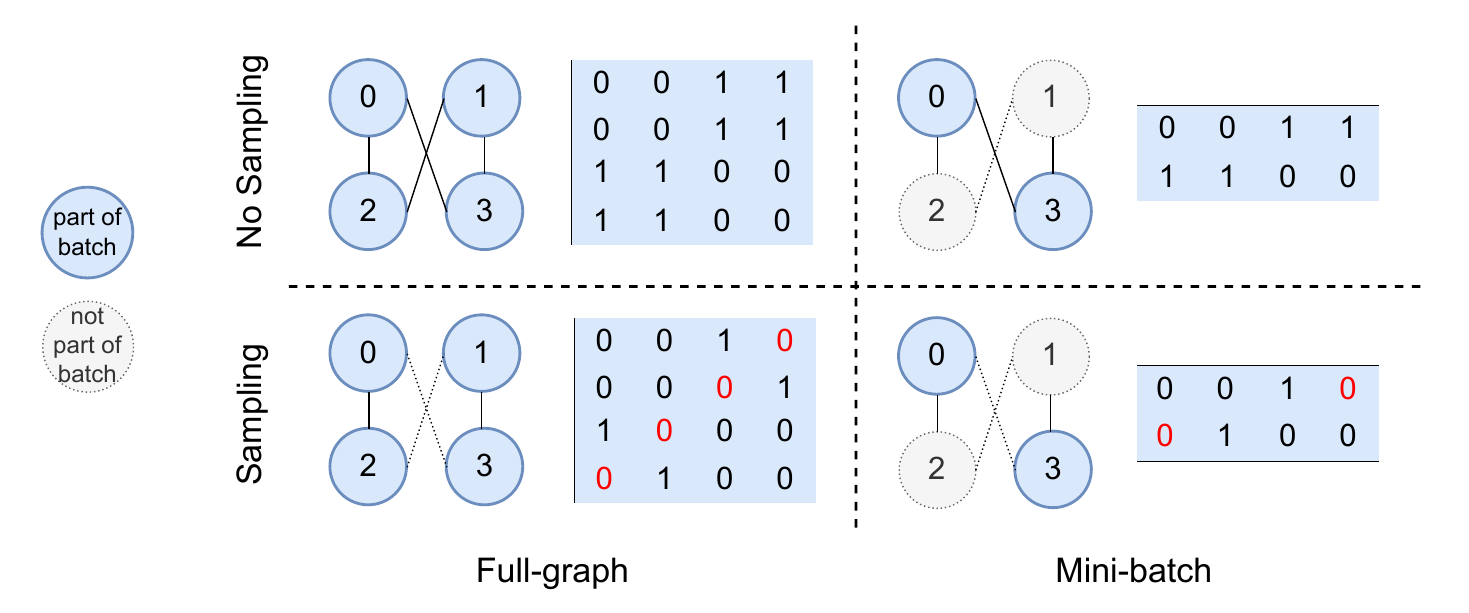}
  \caption{Different paradigms of GNN training that can be combined together,
shown in four quadrants. Each quadrant shows a sample graph and its adjacency
matrix. Blue nodes are part of the batch and grey nodes are not. Solid lines
indicate edges considered during aggregation, and dashed line represent edges
that are not considered. Red values in the adjacency matrix indicate that an
entry has been modified.}
  \label{fig:bg}
\end{figure*}

The edges in a graph are represented by a sparse adjacency matrix $\mathbf{A}
\in \mathbb{R}^{N \times N}$, where $N$ is the number of nodes in the graph.
Prior to training, self-loops are added to $\mathbf{A}$ so that each node's
learned representation includes its own features. $\mathbf{A}$ is then
normalized by scaling each edge $A_{u,v}$ by $\frac{1}{\sqrt{d_u d_v}}$ where
$d_u$ and $d_v$ are the degrees of nodes $u$ and $v$ respectively. This is
common practice to mitigate numerical instabilities such as exploding/vanishing
gradients~\cite{KipfW16}.

The forward pass of a Graph Convolutional Network (GCN) layer $i$ consists of
three key steps:

\vspace{0.08in}
\noindent\textbf{(1)~Aggregation}: Each node has a low-dimensional feature
vector associated with it.  These feature vectors are stored in the features
matrix $\mathbf{F}^{Li} \in \mathbb{R}^{N \times D^{Li}}$ where $D^{Li}$ is the
features dimension at layer $i$.  In the first step of the forward pass, every
node aggregates the features from its immediate neighbors using an aggregation
operator like sum and captures the local graph structure. This is achieved by
performing an SpMM - multiplying the adjacency matrix $\mathbf{A}$ with the
features matrix $\mathbf{F}^{Li} \in \mathbb{R}^{N \times D^{Li}}$. This
results in an intermediate matrix \( \mathbf{H}^{Li} \in \mathbb{R}^{N \times
D^{Li}} \)

\vspace{0.08in}
\begin{equation} \eqnmarkbox[color_1]{a}{\mathbf{H}^{Li}} = \mathrm{SpMM} \left ( \eqnmarkbox[color_2]{b}{\mathbf{A}}, \eqnmarkbox[color_3]{c}{\mathbf{F}^{Li}} \right )
\end{equation}
\annotate[yshift=.5em]{above}{a}{aggregation output}
\annotate[yshift=-.25em]{below,left}{b}{adjacency matrix}
\annotate[yshift=-.25em]{below,right}{c}{features matrix}
\vspace{0.1in}

Without loss of generality, this is shown for the undirected case. For directed
graphs, the adjacency matrix can be transposed for aggregation of features from
incoming neighbors.

\vspace{0.08in}
\noindent\textbf{(2)~Combination}: The aggregated features are transformed into
a new low-dimensional space using a weight matrix \( \mathbf{W}^{Li} \in
\mathbb{R}^{D^{Li} \times D^{Li+1}} \), resulting in an intermediate matrix \(
\mathbf{Q}^{Li} \in \mathbb{R}^{N \times D^{Li+1}} \). 

\vspace{0.15in}
\begin{equation}
    \eqnmarkbox[color_4]{a}{\mathbf{Q}^{Li}} = \mathrm{SGEMM} \left ( \eqnmarkbox[color_1]{b}{\mathbf{H}^{Li}}, \eqnmarkbox[color_6]{c}{\mathbf{W}^{Li}} \right )
\end{equation}
\annotate[yshift=.5em]{above}{a}{combination output}
\annotate[yshift=-.25em]{below,left}{b}{aggregation output}
\annotate[yshift=-.25em]{below,right}{c}{weight matrix}

\vspace{0.08in}
\noindent\textbf{(3)~Activation}: A non-linear activation function $\sigma$
(e.g.  ReLU) is then applied to \( \mathbf{Q}^{Li} \), yielding the output
matrix for the current layer \( \mathbf{F}^{Li + 1} \in \mathbb{R}^{N \times
D^{Li+1}} \).  This will be used as the input to the next layer.

\vspace{0.15in}
\begin{equation}
    \eqnmarkbox[color_3]{a}{\mathbf{F}^{Li + 1}} = \eqnmarkbox[color_5]{b}{\sigma} \left ( \eqnmarkbox[color_4]{c}{\mathbf{Q}^{Li}} \right )
\end{equation}
\annotate[yshift=.5em]{above}{a}{output matrix}
\annotate[yshift=-.25em]{below,left}{b}{activation function}
\annotate[yshift=-.25em]{below,right}{c}{combination output}
\vspace{0.08in}

The corresponding backward pass for layer $i$ involves computing gradients as
follows:

\vspace{0.05in}
\noindent\textbf{(1)}~Compute the gradient of the loss $\mathcal{L}$ with
respect to $\mathbf{Q}^{Li}$:
\begin{equation} \frac{\partial
\mathcal{L}}{\partial\mathbf{Q}^{Li}} = \frac{\partial \mathcal{L}}{\partial
\mathbf{F}^{Li+1}} \eqnmarkbox[color_5]{a}{\odot} \sigma' \left ( \mathbf{Q}^{Li} \right )
\end{equation}
\annotate[yshift=-.25em]{below,left}{a}{element-wise multiplication}
\vspace{0.05in}

\vspace{0.05in}
\noindent\textbf{(2)}~Compute the gradient of the loss with respect to the
weight matrix \( \mathbf{W}^{Li} \):
\begin{equation}
    \frac{\partial \mathcal{L}}{\partial \mathbf{W}^{Li}} =
    \mathrm{SGEMM} \left ( \left ( \mathbf{H}^{Li} \right )^\top, \frac{\partial \mathcal{L}}{\partial
    \mathbf{Q}^{Li}} \right )
\end{equation}

\vspace{0.05in}
\noindent\textbf{(3)}~Compute the gradient of the loss with respect to \(
\mathbf{H}^{Li} \):
\begin{equation}
    \frac{\partial \mathcal{L}}{\partial \mathbf{H}^{Li}} = \mathrm{SGEMM} \left ( \frac{\partial
    \mathcal{L}}{\partial \mathbf{Q}^{Li}}, \left ( \mathbf{W}^{Li} \right )^\top \right )
\end{equation}

\vspace{0.05in}
\noindent\textbf{(4)}~Compute the gradient of the loss with respect to \(
\mathbf{F}^{Li} \):
\begin{equation}
    \frac{\partial \mathcal{L}}{\partial \mathbf{F}^{Li}} = \mathrm{SpMM} \left ( \mathbf{A}^\top, \frac{\partial \mathcal{L}}{\partial \mathbf{H}^{Li}} \right )
\end{equation}

The gradient $\frac{\partial \mathcal{L}}{\partial \mathbf{F}^{L0}}$ at the
first layer is then used to update the input features and learn meaningful node
embeddings.

\subsection{Different Paradigms of GNN Training}

Four main GNN training paradigms exist (see Figure~\ref{fig:bg}).
\textit{Full-graph} training (upper-left) uses the entire graph in each
iteration, updating all node features and requiring the entire graph in memory.
This makes no approximations but is memory-intensive.  \textit{Mini-batch}
training (upper-right) updates only a small subset of nodes per iteration
(e.g., Nodes 0 and 3), but suffers from neighborhood explosion in deeper
GNNs~\cite{chen2018stochastictraininggraphconvolutional}. To address this,
\textit{Mini-batch sampling} (bottom-right), the most common paradigm, combines
mini-batching with neighbor sampling at each layer and only uses some edges for
aggregation.  Finally, \textit{Full-graph sampling} (bottom-left) uses the
entire graph as a batch but samples edges, which is less common.

While there are some sampling algorithms that have fairly successful adoption,
they still lack a community standard.  GraphSAGE~\cite{graphsage} samples a
fixed number of neighbors per node, while FastGCN~\cite{fastgcn} samples per
layer.  LADIES enhances FastGCN by considering inter-layer
dependencies~\cite{ladies}. Cluster-GCN~\cite{cluster-gcn} samples within dense
subgraphs.  Recent work explores adaptive sampling (GRAPES~\cite{grapes}) and
handling homophilic/heterophilic graphs (AGS-GNN~\cite{ags-gnn}).  However,
sampling introduces a trade-off between accuracy and efficiency, causing bias
and variance~\cite{liu2021samplingmethodsefficienttraining} in training. The
limited scale of graphs used in these studies (max of 2.5 million nodes) raises
concerns about information loss on larger, real-world datasets with different
structural properties. Consequently, the effectiveness of sampling remains
inconclusive, motivating our focus on distributed full-graph training.

\begin{figure*}[t]
  \centering
  \includegraphics[width=\textwidth]{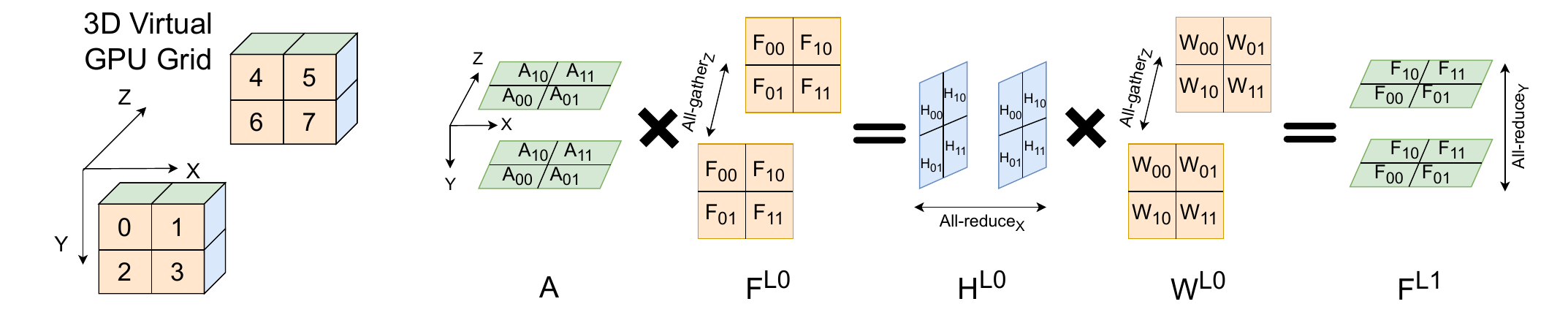}
  \caption{An overview of the 3D tensor parallel algorithm for GNN training.
Eight GPUs are arranged in a 3D grid (X=Y=Z=2) and matrices in layer 0 of the
network are distributed across different planes (shown in different
colors).}
  \label{fig:method}
\end{figure*}

\subsection{Distributed Full-graph GNN Training}
\label{sec:relatedwork}

Early distributed full-graph GNN training frameworks include
ROC~\cite{MLSYS2020ROC}, which partitions graphs using online linear regression
and balances CPU-GPU transfer with GPU memory. CAGNET~\cite{cagnet} uses
tensor-parallel algorithms (1D, 1.5D, 2D, 3D) for SpMM. While the 2D and 3D
algorithms offer asymptotic communication reduction, the 1D and 1.5D algorithms
scale better due to lower constants. A sparsity-aware version of CAGNET's
1D/1.5D algorithms~\cite{cagnet-sparsity-aware} improves performance by
communicating only necessary features.  MG-GCN~\cite{mg-gcn} optimizes CAGNET
with communication-computation overlap.  RDM~\cite{rdm} builds on CAGNET with
near communication-free training by replicating one of the matrices.

Other full-graph frameworks introduce approximations for scalability.
BNS-GCN~\cite{bns-gcn} partitions with METIS and samples boundary nodes, but
its convergence on diverse datasets needs further validation.
PipeGCN~\cite{pipegcn} pipelines communication and computation, potentially
causing stale features/gradients, with sensitivity varying across graphs.
DGCL~\cite{dgcl} minimizes communication using graph characteristics and
cluster topology. NeutronTP~\cite{neutrontp} uses tensor parallelism by only
distributing the features to avoid load imbalance.

\begin{table}[h]
    \caption{Summary of state of the art in distributed full-graph GNN
training. The number of nodes and edges of the graph datasets, and number of
GPUs are the largest values reported in each paper.}
    \label{tab:related-work}
    \begin{tabular}{lcccc}
        \toprule
        Name & Year & \# Nodes & \# Edges & \# GPUs \\
        \midrule
        AdaQP~\cite{adaqp} & 2023 & 2.5M & 114M & 8 \\
        RDM~\cite{rdm} & 2023 & 3M & 117M & 8 \\
        MG-GCN~\cite{mg-gcn} & 2022 & 111M & 1.6B & 8 \\
        Sancus~\cite{sancus} & 2022 & 111M & 1.6B & 8 \\
        MGG~\cite{mgg} & 2023 & 111M & 1.6B & 8 \\
        DGCL~\cite{dgcl} & 2021 & 3M & 117M & 16 \\
        ROC~\cite{MLSYS2020ROC} & 2020 & 9.5M & 232M & 16 \\
        NeutronStar~\cite{neutronstar} & 2022 & 42M & 1.5B & 16 \\
        GraNNDis~\cite{granndis} & 2024 & 111M & 1.6B & 16 \\
        NeutronTP~\cite{neutrontp} & 2024 & 244M & 1.7B & 16 \\
        CDFGNN~\cite{cdfgnn} & 2024 & 111M & 1.8B & 16 \\
        PipeGCN~\cite{pipegcn} & 2022 & 111M & 1.6B & 32 \\
        CAGNET~\cite{cagnet} & 2020 & 14.2M & 231M & 125 \\
        BNS-GCN~\cite{bns-gcn} & 2022 & 111M & 1.6B & 192 \\
        SA+GVB~\cite{cagnet-sparsity-aware} & 2024 & 111M & 1.6B & 256 \\
        \midrule
        \plexus (this work) & 2025 & 111M & 1.6B & 2048 \\
        \bottomrule
    \end{tabular}
\end{table}

Table~\ref{tab:related-work} shows limited scaling across many GPUs in existing
full-graph works, with a handful using more than 16 GPUs. Many focus on 1D SpMM
variants, lacking a practical scalable 3D algorithm despite its theoretical
communication advantages. This motivates \plexus, our framework aiming for
approximation-free, scalable 3D full-graph training for large graphs and high
GPU counts.

\section{A Three-dimensional Tensor Parallel Approach to Full-graph GNN Training}
\label{sec:method}
We now describe our approach to parallelizing a GCN layer and the entire
network, and our adaptation of Agarwal's 3D parallel matrix multiply algorithm
for GNN training in \plexus.

\subsection{Parallelizing a Single GCN Layer}
\label{sec:3d-algo}

Tensor parallelism is a popular strategy for parallelizing GNN training. While
previous works have experimented with 1D to 3D tensor parallel approaches, in
this work, we focus on 3D tensor parallelism.  We take inspiration from Agarwal
et al.'s three-dimensional (3D) parallel matrix multiplication
algorithm~\cite{agarwal-3d} for distributing matrices and parallelizing matrix
multiplication kernels across multiple GPUs.  Below, we describe how we adapt
this 3D matrix multiplication approach to parallelize GNN training and Sparse
Matrix-Matrix Multiplication (SpMM) computations.

Given a number of GPUs, $G$, in a job allocation, we first arrange the GPUs
into a 3D virtual grid. We refer to the number of GPUs along each dimension as
\( G_x \), \( G_y \), and \( G_z \) respectively, such that $G = G_x \times G_y
\times G_z$. Each GPU creates process groups that allow it to communicate with
its neighbors in each of the three dimensions of the grid.  The matrices in a
layer are then distributed across this grid. Here, we describe how this is done
for the first layer of the GCN, and this can be applied similarly to the other
layers.

\begin{figure}[h]
  \centering
  \includegraphics[width=2.5in]{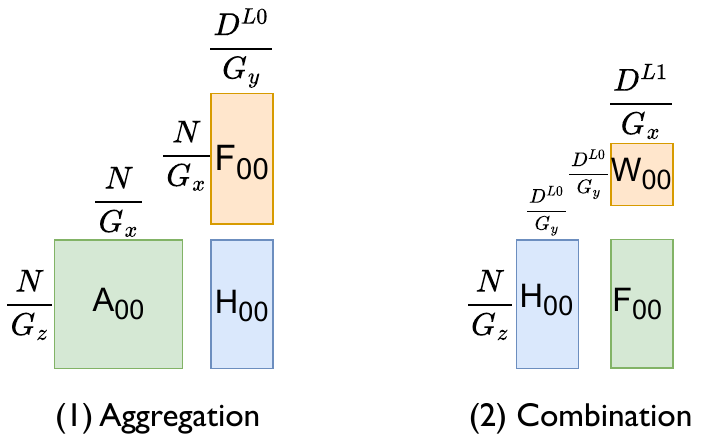}
  \caption{Shapes of the matrix shards (sub-blocks) in the first layer on a single GPU,
showing two key matrix multiplications in the forward pass.}
  \label{fig:shapes}
\end{figure}

First, we shard (divide and map to different GPUs) the sparse adjacency matrix,
$A$, across the \( ZX \)-plane and replicate it across the \( Y \)-parallel
process group (see Figure~\ref{fig:method}). Then we shard the input features
matrix, $F^{L0}$, across the \( XY \)-plane and further shard it across the \(
Z \)-parallel process group.  The reason that $F^{L0}$ is sharded and not
replicated across the third process group is to save memory. Since the input
features are made trainable to learn node embeddings, they have gradients and
optimizer states associated with them which are additional memory requirements.
Finally, we shard the weights across the \( YX \)-plane and also further across
the \( Z \)-parallel process group due to the additional memory requirements of
the gradients and optimizer states.  Figure~\ref{fig:shapes} shows the shapes
of the matrix shards (sub-blocks or sub-matrices) for layer 0.

\begin{figure*}[t]
  \centering
  \includegraphics[width=\textwidth]{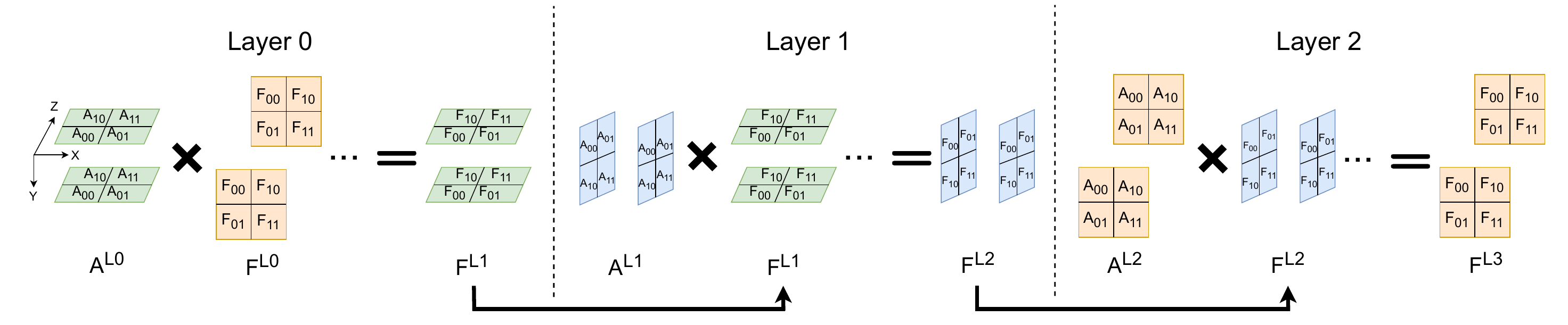}
  \caption{Applying the 3D tensor parallel algorithm to all layers of a
3-layer GCN, connecting the output of one layer to the input of the next using
unique shards of the adjacency matrix.}
  \label{fig:method-layer}
\end{figure*}

Pseudo code for the forward and backward pass of layer 0 is shown in
Algorithm~\ref{alg:tensor-parallel} and~\ref{alg:tensor-parallel1}
respectively. Before describing the algorithm, note that when we refer to any
matrix, it is a shard of that matrix on a given GPU.  \highlight{soul_1}{Lines
3-5} show the aggregation step, in which the input features matrix shard $F$ is
all-gathered across the \( Z \)-parallel process group since it is additionally
sharded across this dimension of the grid. The adjacency matrix shard $A$ is
then multiplied with $F$ to get the aggregation output $H$. Since this results
in a partial output, an all-reduce is performed on $H$ across the \( X
\)-parallel process group.

\begin{algorithm}[h]
  \caption{Forward Pass of Layer 0}
  \label{alg:tensor-parallel}
  \begin{algorithmic}[1]
  \Function{Forward}{$\mathbf{A}, \mathbf{F}, \mathbf{W}$}
  \sethlcolor{soul_1}
  \State // Step 1: Aggregation
  \State \hl{\textbf{All-gather} $\mathbf{F}$ across \textbf{Z}-parallel group}
  \State \hl{$\mathbf{H} = \textbf{SpMM}(\mathbf{A}, \mathbf{F})$}
  \State \hl{\textbf{All-reduce} $\mathbf{H}$ across \textbf{X}-parallel group}
  \Statex

  \sethlcolor{soul_2}
  \State // Step 2: Combination
  \State \hl{\textbf{All-gather} $\mathbf{W}$ across \textbf{Z}-parallel group}
  \State \hl{$\mathbf{Q} = \textbf{SGEMM}(\mathbf{H}, \mathbf{W})$}
  \State \hl{\textbf{All-reduce} $\mathbf{Q}$ across \textbf{Y}-parallel group}
  \Statex

  \sethlcolor{soul_3}
  \State // Step 3: Non-linear Activation
  \State \hl{$\mathbf{F} = \bm{\sigma}(\mathbf{Q})$}
  \State \hl{\textbf{Return} $\mathbf{F}$}
  \EndFunction
  \end{algorithmic}
\end{algorithm}

\highlight{soul_2}{Lines 7-9} show the next combination step. First, the
weights matrix shard $W$ is all-gathered across the \( Z \)-parallel process
group since it is additionally sharded across this dimension of the grid.  The
intermediate output from the aggregation is then multiplied by the weights
matrix. This again results in a partial output $Q$, which is all-reduced across
the \( Y \)-parallel process group. Finally, we apply a non-linear activation
on this and return it to be used in the next layer (\highlight{soul_3}{lines
11-12}). These series of matrix multiplications and all-reduce steps are also
demonstrated visually in Figure~\ref{fig:method}. The backward pass for the
first layer is shown in Algorithm~\ref{alg:tensor-parallel1}.

\begin{algorithm}[t]
  \caption{Backward Pass of Layer 0}
  \label{alg:tensor-parallel1}
  \begin{algorithmic}[1]
  \Function{Backward}{$\frac{\partial \mathcal{L}}{\partial Q}$}
  \State \ensuremath{\frac{\partial \mathcal{L}}{\partial W} = \textbf{SGEMM}(H^T, \frac{\partial \mathcal{L}}{\partial Q})}
  \State \textbf{Reduce-scatter} \ensuremath{\frac{\partial \mathcal{L}}{\partial W}} across \textbf{Z}-parallel group
  \Statex

  \State \textbf{All-gather} $\mathbf{W}$ across \textbf{Z}-parallel group
  \State \ensuremath{\frac{\partial \mathcal{L}}{\partial H} = \textbf{SGEMM}(\frac{\partial \mathcal{L}}{\partial Q}, W^T)} 
  \State \textbf{All-reduce} \ensuremath{\frac{\partial \mathcal{L}}{\partial H}} across \textbf{X}-parallel group
  \Statex

  \State \ensuremath{\frac{\partial \mathcal{L}}{\partial F} = \textbf{SpMM}(A^T, \frac{\partial \mathcal{L}}{\partial H})}

  \State \textbf{Reduce-scatter} \ensuremath{\frac{\partial \mathcal{L}}{\partial F}} across \textbf{Z}-parallel group
  \Statex

  \State \textbf{Return} \ensuremath{\frac{\partial \mathcal{L}}{\partial F}, \frac{\partial \mathcal{L}}{\partial W}}
  \EndFunction
  \end{algorithmic}
\end{algorithm}

\subsection{Parallelizing All Layers in the Network}
\label{sec:entire-gnn}

The parallelization of other layers in the GNN is similar to the first layer
but we need to address a subtle but important detail first.  As can be seen in
Figure~\ref{fig:method}, the output of the first layer \( F^{L1} \) is sharded
across the \( ZX \)-plane. However, this will also be the input to the next
layer, which becomes an issue since the adjacency matrix \( A \) of the next
layer is also sharded across the \( ZX \)-plane, and so the dimensions of the
two matrices are incompatible.  To resolve this, we either need to communicate
\( F^{L1} \) to the \( XY \)-plane or communicate \( A \) to the \( YZ
\)-plane. Unfortunately, these solutions would add increased communication
complexity and are non-trivial to implement efficiently.

To address this problem, we store a separate shard of the adjacency matrix \(
A^{L1} \) that is sharded across the \( YZ \)-plane for the next layer \( L1
\). Similarly, for the third layer \( L2 \), we store a shard of the adjacency
matrix \( A^{L2} \) that is sharded across the \( XY \)-plane. This ensures
that the dimensions of the matrices are compatible for local computations.
This scheme is shown in Figure~\ref{fig:method-layer}, where we can see how the
three adjacency matrix shards allow for the output of one layer to be used as
the input for the next layer. Importantly, this does not result in needing more
than three unique shards of the adjacency matrix.  The output of the third
layer \( F^{L3} \) is sharded across the \( XY \)-plane, which is the same
plane that \( F^{L0} \) is sharded across. So for the fourth layer \( L3 \), we
can now reuse \( A^{L0} \) and then repeat using the same adjacency matrix
shards for subsequent layers.

This process of cycling through three different adjacency shards for different
layers also changes a few communication steps in
Algorithm~\ref{alg:tensor-parallel}.  For subsequent layers after the first
one, the features matrix \( F \) will only be sharded across two dimensions of
the grid since it does not have optimizer states like the input features. This
means that the first all-gather in the forward pass (line 2) will not take
place. Likewise, the last reduce-scatter (line 8) in the backward pass is
changed to an all-reduce since the gradients are replicated across the third
process group.  Using different shards of the adjacency matrix is the main
change to parallelize all the layers of the model and the core idea remains the
same.

\section{Performance Model}
Next, we describe the performance model we have developed to identify
near-optimal 3D configurations of the virtual GPU grid. We model both the SpMM
computation and communication times.

\subsection{Modeling Computation}

\plexus shards matrices such that local matrix operations should take the same
amount of time across different 3D configurations, assuming a uniform
distribution of nonzeros. We show this in the derivation below, where we see
that the total number of FLOPs needed to calculate the aggregation output \(H\)
is a term that is constant across all configurations for $G = G_x \times G_y \times G_z$ GPUs.

Given the number of nodes in the graph \( N \) and the input features dimension \( D^{L0} \), the number of elements in \( H \) in the first layer is:
    \vspace{0.2in}
    \begin{equation}
    \frac{\eqnmarkbox[color_1]{a}{N}}{G_z} \times \frac{\eqnmarkbox[color_3]{c}{D^{L0}}}{G_y}
    \end{equation}
    \annotate[yshift=.5em]{above,left}{a}{number of nodes}
    \annotate[yshift=.5em]{above,right}{c}{input features dimension}

Given the number of nonzeros in the adjacency matrix \( \mathit{NNZ} \), the number of floating point operations per element is:
    \vspace{0.2in}
    \begin{equation}
    \mathcal{O} \left(\frac{2 \times \eqnmarkbox[color_2]{e}{\mathit{NNZ}}}{N \times G_x} \right)
    \end{equation}
    \annotate[yshift=.5em]{above,left}{e}{number of nonzeros}

Hence, the total number of floating point operations to calculate the
aggregation output \(H\) is a result of multiplying the expressions in equations (4.1) and
(4.2) together:
    \begin{equation}
    \mathcal{O} \left(\frac{2 \times \mathit{NNZ} \times D^{L0}}{\eqnmarkbox[color_4]{j}{G}} \right)
    \end{equation}
    \annotate[yshift=-.25em]{below,left}{j}{number of GPUs}
    \vspace{0.2in}

Despite expecting similar computation times for different 3D configurations, in
practice, we observe that SpMM times vary across configurations.  We
hypothesize that shorter-fatter dense matrices lead to more efficient SpMMs.
This is consistent with the literature optimizing tall-skinny dense SpMM.  Yang
et al.~\cite{sparse-design-principles} propose row-splitting for coalesced
memory access, which they note is more efficient with fewer nonzeros per row.
This is achieved by configurations in our algorithm reducing the common
dimension of local multiplications.  Selvitopi et
al.~\cite{distributed-memory-spmm} show non-ideal scaling of SpMM time with the
number of processors and that the algorithm choice can impact scaling.

To test our hypothesis, we took the adjacency and feature matrices from
ogbn-products and multiplied them under two different configurations for 64
GPUs. In config U, $G_x=64$ and the common dimension is sharded by 64, reducing
the number of nonzeros per row. In config V, $G_y=64$ and the columns of the
dense matrix are sharded by 64, making it skinny. Both of these have the same
workload in terms of the number of FLOPs.  However, we observed that V was
\tweakedsim 8$\times$ slower. After profiling with Nsight Compute~\cite{ncu} (metrics in
Table~\ref{tab:comp-model}), we noticed that it launched \tweakedsim 64 times more
blocks, which is proportional to its 64$\times$ larger common dimension size. This
means less work per block and a higher number of smaller memory requests.
Consequently, V's L2 Cache and DRAM throughput were drastically lower, and
uncoalesced global memory accesses were much higher, indicating poor memory
access patterns and suboptimal memory utilization in the tall-skinny dense SpMM
regime.

\begin{table}[h]
  \caption{Nsight Compute metrics for \texttt{SpMM(A, H)} on a single GPU for two configurations of Plexus -- U ($G_z=1$, $G_x=64$, $G_y=1$) and V ($G_z=1$, $G_x=1$, $G_y=64$).}
  \label{tab:comp-model}
  \centering
  \resizebox{\columnwidth}{!}{
  \begin{tabular}{lrr}
      \toprule
      Metric & U & V \\
      \midrule
      Grid Size        & 20,223 & 1,313,241 \\
      Uncoalesced Global Memory Access Sectors       & 84,960 & 3,939,912 \\
      L2 Cache Throughput        & 61.31 & 12.65 \\
      DRAM Throughput   & 72.83 & 8.24 \\
      \bottomrule
  \end{tabular}
  }
\end{table}

In \plexus, we introduce a computational model to predict which configurations
result in more efficient SpMMs. The model is shown for the first layer using
the equations below:
\begin{align*}
  \mathrm{flops\_cost} &= \mathit{NNZ} \times D^{L0} \\
  \text{fwd\_penalty} &= \frac{N}{G_x} \times \frac{G_y}{D^{L0}} \\
  \text{bwd\_penalty} &= \frac{N}{G_z} \times \frac{G_y}{D^{L0}}
\end{align*}
\begin{align}
  \text{comp\_cost} & = \sqrt{\text{flops\_cost}} \\ \nonumber
                    & \times \left(1 + \text{fwd\_penalty} + \text{bwd\_penalty}\right)
\end{align}

The first term \(\mathrm{flops\_cost}\) is proportional to the total FLOPs,
which is the number of nonzeros \( \mathit{NNZ} \) in the sparse matrix \( A \)
multiplied by the number of columns \( D^{L0} \) in the dense matrix \( F \).
The second term \(\mathrm{fwd\_penalty}\) ranks certain configurations as
better than others based on the matrix shape. This term is first weighted
proportional to the size of the matrix \( F \)'s first dimension:
\(N/G_{\text{x}}\) (the common
dimension). It is then weighted inversely proportional to the size of
the second dimension of \( F \): \(D^{L0}/G_{\text{y}}\). This penalizes configurations causing tall-skinny dense matrices.
A similar calculation is done for the backward pass SpMM.

The final computational cost is calculated as the square root of \( \mathrm{flops\_cost}
\) (to reduce outlier impact of larger matrices), multiplied by penalty terms
to account for poor matrix shapes, and summed across all layers.  To convert
this to time, we performed runs on Perlmutter across various datasets,
configurations, and GPU counts (including all ogbn-products configurations on
64 GPUs).  We then used scikit-learn~\cite{scikit-learn} to fit a linear
regression model to these 67 data points, determining coefficients for our
three terms to predict SpMM time for any configuration.

To validate our model, we used a random train-test split of 70-30 for 1000
independent iterations.  We recorded an average $R^2$ of 0.89 and $RMSE$ of
16.8 ms for the train splits, and an average $R^2$ value of 0.79 and $RMSE$ of
20.1 ms for the test splits, indicating that the model is able to predict the
SpMM time with a relatively high degree of accuracy and can generalize fairly
well. The learned coefficients for the three terms are approximately $7.8
\times 10^{-4}$, $7.8 \times 10^{-10}$, and $-2.6 \times 10^{-10}$.

\subsection{Modeling Communication}

Different 3D grid configurations significantly impact communication time and
overall performance, especially at scale.  Optimal configuration selection is
non-trivial. Several works model communication time for distributed deep
learning, such as ATP~\cite{cheng2023atp}, Alpa~\cite{alpa},
AxoNN~\cite{singh:ipdps2022, singh:sc2024}, Oases~\cite{li2023automated-oases},
and DGCL~\cite{dgcl}. \plexus adapts AxoNN's communication
model~\cite{singh:sc2024}, which uses ring algorithm equations from Thakur et
al.~\cite{thakurimproving2003} and
Rabenseifner~\cite{rabenseifneroptimization2004}. The latency term is omitted
since the messages are large and bandwidth-bound.  The all-reduce time for a
buffer of size $M$ across $G$ GPUs with bandwidth $\beta$ can be modeled as:
\vspace{0.05in}
\begin{equation}
\eqnmarkbox[color_1]{a}{T_{\mathrm{all-reduce}}} = \frac{2}{\eqnmarkbox[color_2]{b}{\beta}} \times \left(\frac{\eqnmarkbox[color_4]{c}{G} - 1}{G}\right) \times \eqnmarkbox[color_6]{d}{M}
\end{equation}
\annotate[yshift=.5em]{above,left}{a}{time}
\annotate[yshift=.5em]{above,right}{c}{number of GPUs}
\annotate[yshift=-.25em]{below,left}{b}{bandwidth}
\annotate[yshift=-.25em]{below,right}{d}{buffer size}
\vspace{0.1in}

\plexus extends this across layers by using the appropriate matrix dimensions
and process group sizes for each layer, as described in
Section~\ref{sec:entire-gnn}.  The model considers GPU topology, prioritizing
\(Y\), \(X\), and then \(Z\) parallelism within a node.  If a process group is within
a node, it can utilize intra-node bandwidth $\beta_{\mathrm{intra}}$.  Otherwise, it
is bound by inter-node bandwidth $\beta_{\mathrm{inter}}$, which can potentially be lower
due to link contention.  We show how this is calculated for $\beta_z$, bandwidth along the $Z$-parallel group, in the
following equation:
\begin{equation}
  \beta_{z} =
  \begin{cases}
      \beta_{\text{intra}} & \text{if } G_{\text{x}} \times G_{\text{y}} \times G_{\text{z}} \leq G_{\text{node}} \\
      \frac{\beta_{\text{inter}}}{\min \left ( G_{\text{node}}, G_{\text{x}} \times G_{\text{y}} \right )} & \text{otherwise}
  \end{cases}
\end{equation}
where $G_{\text{node}}$ is the number of GPUs within a node.

After the effective bandwidths are similarly calculated for \( \beta_{x} \) and
\( \beta_{y} \), we can plug them in to the equations for each collective and
calculate the predicted communication times for each configuration.

\subsection{Unified Performance Model}

\begin{figure}[h]
  \centering
  \includegraphics[width=0.9\columnwidth]{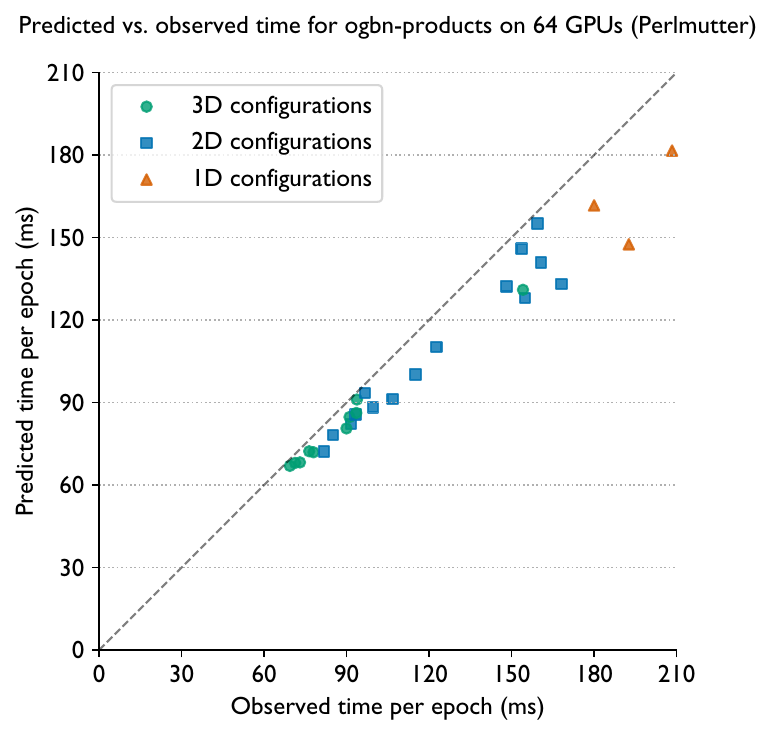}
  \caption{Validating the performance model for the ogbn-products dataset on 64 GPUs of Perlmutter.}
 \label{fig:perf-model}
\end{figure}

We combine predicted SpMM time and communication time to estimate total epoch
time for each configuration, neglecting smaller dense computation and loss
calculation. Figure~\ref{fig:perf-model} shows results for ogbn-products on 64
Perlmutter GPUs, indicating better performance for 3D configurations over 2D
and 1D. The three-layer GCN favors symmetric configurations for balanced
communication and SpMM efficiency. As we can observe, a strong correlation exists between
predicted and observed epoch times, accurately predicting top configurations.

\section{Performance Optimizations}
\label{sec:opt}
Parallelizing graph neural networks can pose unique challenges in the form of
load imbalance caused by uneven sparsity patterns and high communication
overheads arising due to the extremely large sizes of graphs. We address some
of these issues in \plexus by introducing several optimizations that improve
the performance of our framework.

\subsection{Double Permutation for Load Balancing}

The sparse and uneven distribution of nonzeros in the adjacency matrix can
cause load imbalance among matrix shards assigned to different GPUs, leading to
computational stragglers and slower training. Graph partitioners such as
METIS~\cite{metis} can be used to partition graphs to minimize edge cuts and
balance vertices, which is beneficial for fine-grained communication. However,
the all-reduce in \plexus is performed on dense aggregation outputs and does
not require graph structure awareness for communication. While graph
partitioners distribute rows/nodes, our 2D matrix decomposition requires
even nonzeros to be evenly distribution across 2D shards.

Node permutation offers a simple solution without complex optimization or graph
structure knowledge. Unlike graph partitioning, which requires re-partitioning
for different GPU counts, permutation is a one-time preprocessing step for each
graph dataset. The na\"ive permutation scheme uses a permutation matrix \(P\) to
map original node indices to permuted indices.

\begin{equation}
\vspace{0.15in}
\eqnmarkbox[color_3]{e}{F^{L1}} = \sigma \left ( \left ( \eqnmarkbox[color_1]{d}{P} \eqnmarkbox[color_2]{a}{A} P^T \right ) \left ( P \eqnmarkbox[color_3]{b}{F^{L0}} \right ) \eqnmarkbox[color_6]{c}{W^{L0}} \right )
\end{equation}
\annotate[yshift=.5em]{above,left}{e}{output features}
\annotate[yshift=.5em]{above,right}{a}{adjacency matrix}
\annotate[xshift=-1em,yshift=-.25em]{below,left}{d}{permutation matrix}
\annotate[yshift=-.25em]{below,left}{b}{input features}
\annotate[yshift=-.25em]{below,right}{c}{weight matrix}
\vspace{0.05in}
\begin{equation}
F^{Li} = \sigma \left ( \left ( P A P^T \right ) F^{Li-1} W^{Li-1} \right)
\end{equation}

Equation (5.1) is used for the first layer and (5.2) is used for all subsequent layers.
The same permutation is applied to adjacency matrix columns and input features
rows to maintain output consistency for subsequent layers. This preprocessing
step significantly reduces load imbalance.  However, due to dense graph
clusters, a single permutation is insufficient, as nonzeros remain concentrated
around diagonal blocks. To further disrupt community coupling, we apply
distinct permutations ($P_r$ for rows, $P_c$ for columns) to the adjacency
matrix, repeating this two-permutation scheme every two layers for more
effective nonzero redistribution.  This requires storing two adjacency matrix
versions.
\vspace{0.2in}
\begin{equation}
F^{L1} = \sigma \left ( \left ( \eqnmarkbox[color_1]{a}{P_r} A \eqnmarkbox[color_5]{b}{P_c^T} \right ) \left (P_c F^{L0} \right ) W^{L0} \right )
\end{equation}
\annotate[yshift=.5em]{above,left}{a}{permutation matrix for rows}
\annotate[yshift=.5em]{above,right}{b}{permutation matrix for columns}
\begin{equation}
F^{Li} = \sigma \left ( \left ( P_c A P_r^T \right ) F^{Li-1} W^{Li-1} \right )
\end{equation}

Alternating between two permutations ($P_r$ and $P_c$) further balances
computation by disrupting tightly coupled communities.
Table~\ref{tab:permutation_comparison} shows near-perfect load balance on the
europe\_osm dataset (8x8 shards) with double permutation, outperforming the na\"ive single
permutation.

\begin{table}[h]
  \caption{Comparison of different permutation methods, showing the ratio of the maximum
  number of non-zeros to the mean across 8x8 shards of the adjacency matrix for the europe\_osm dataset.}
  \label{tab:permutation_comparison}
  \centering
  \begin{tabular}{lcc}
  \hline Method & Max/Mean \\ \hline
  Original  & 7.70 \\ Single permutation & 3.24
  \\ Double permutation (this work)  & 1.001 \\ \hline
  \end{tabular}
\end{table}

Adopting this optimization increases the memory required to store each
adjacency matrix shard by a factor of two. Since the number of such shards is $\min(3, L)$
for $L$ GCN layers, the memory overhead of storing the shards after applying
this optimization then becomes $\min(6, L)$. Given that the number of
GCN layers is typically small (two to four) to avoid oversmoothing~\cite{deeper-insights-gnn}, this is a
reasonable trade-off for improved load balance and performance.

\subsection{Blocked Aggregation}

\begin{figure}[h]
  \centering
  \includegraphics[width=0.49\columnwidth]{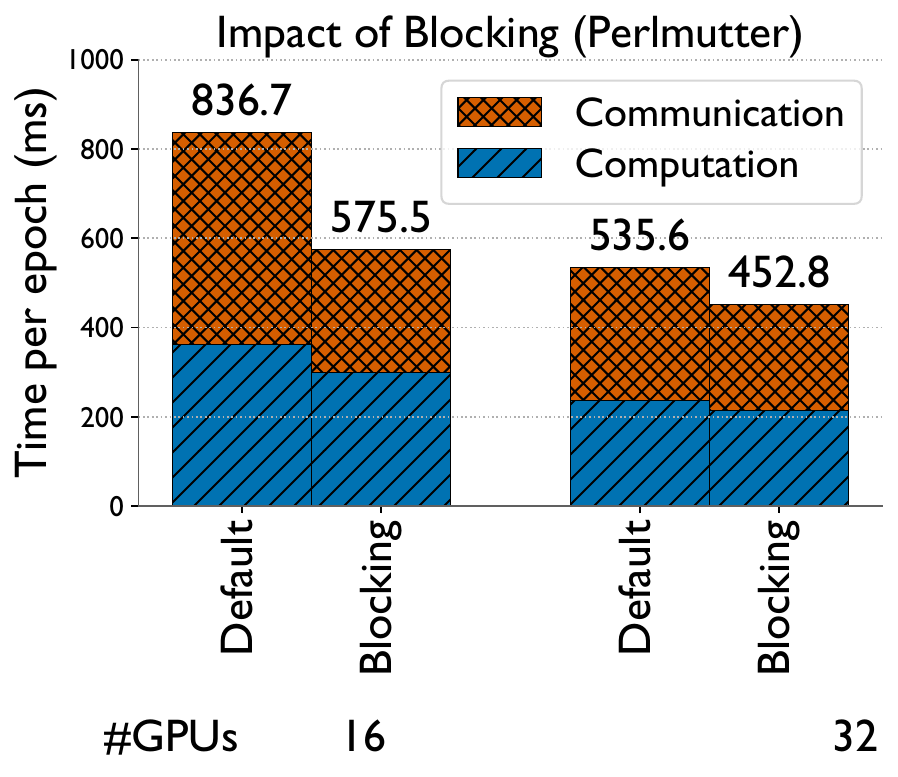}
  \includegraphics[width=0.49\columnwidth]{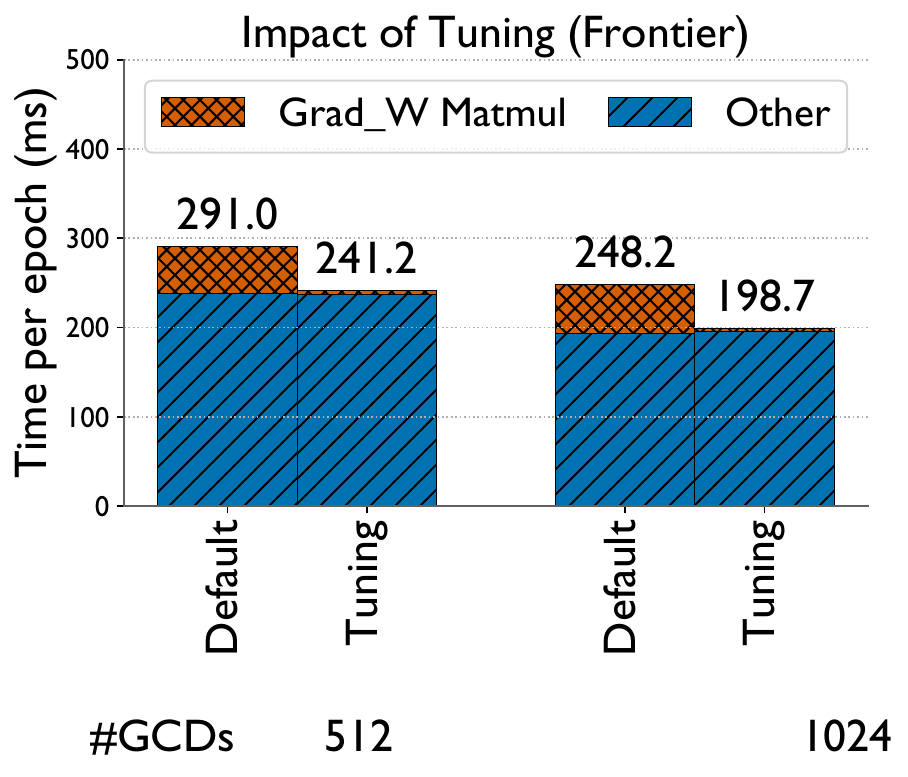}
  \caption{Impact of blocked aggregation on performance for Isolate-3-8M on 16
and 32 GPUs of Perlmutter (left). Impact of dense matrix multiplication tuning
on performance for products-14M on 512 and 1024 GCDs of Frontier (right).}
  \label{fig:blocking}
\end{figure}

While our double permutation achieves near-perfect adjacency matrix load
balance, we observed performance variability in the forward pass SpMM across epochs on
larger datasets (Isolate-3-8M, products-14M) for a modest number of GPUs (8-32). This leads to
load imbalance in the subsequent all-reduce and increased average epoch
time. To address this, we optimized the aggregation by blocking the sparse
adjacency matrix into smaller row-blocks, since we did not observe this for
smaller matrices. After each block's SpMM, an all-reduce is performed on it,
and blocks are concatenated at the end. This mitigated performance variability in the SpMM,
and significantly reduced communication also as a side effect, as shown in
Figure~\ref{fig:blocking} (left).

\subsection{Dense Matrix Multiplication Tuning}

Despite dense matrix multiplication
taking a small amount of time in our workloads, we observed scaling issues on Frontier at high GCD
counts (>= 512 GCDs) with large datasets (such as Isolate-3-8M and products-14M) for the
$\frac{\partial \mathcal{L}}{\partial W}$ calculation, where the first matrix
was transposed.
GEMM BLAS kernels have NN, NT, TN, TT modes with varying performance (e.g., NT and
TN can be slower~\cite{shi2017tnvnn}). We optimized these dense kernels by reversing the multiplication order:
$\frac{\partial \mathcal{L}}{\partial W} = \left ( \mathrm{SGEMM} \left ( {\frac{\partial
\mathcal{L}}{\partial Q}}^T, H \right ) \right )^T$.  Figure~\ref{fig:blocking} (right) shows a
significant time reduction for this GEMM on Isolate-3-8M (from \tweakedsim 50 ms to
negligible), enabling \plexus to scale to 1024 GCDs on Frontier.

\subsection{Parallel Data Loading}

Many GNN frameworks load entire datasets into CPU memory before transferring
shards to the GPU, which is unsustainable for large graphs. \plexus implements
a parallel data loader to avoid this.  It shards processed data into 2D files
offline (e.g., 8x8), and the data loader for each GPU only loads, merges, and extracts the
shards it needs. This significantly reduces CPU memory usage and data loading time. For
ogbn-papers100M on 64 GPUs, CPU memory requirements decreased from 146 GB to 9
GB (16x16 shards), and loading time from 139s to 7s with parallel data loading.

\section{Experimental Setup}
\label{sec:setup}
Below, we provide details of the experimental setup used to evaluate \plexus,
including the supercomputer platforms and datasets used, model details, and
other state-of-the-art (SOTA) frameworks we compare against.

\subsection{Details of Supercomputer Platforms}

Our experiments were conducted on Perlmutter at NERSC, Lawrence Berkeley
National Laboratory, and Frontier at OLCF, Oak Ridge National Laboratory. The
GPU partition of Perlmutter is connected by the HPE Slingshot 11 network, and
has two kinds of compute nodes. 1,536 nodes have four NVIDIA A100 GPUs each with 40 GB
of HBM2 memory per GPU. 256 additional nodes have four A100 GPUs with 80 GB of
HBM2 memory per node.  We use the 80 GB nodes for runs on 64 and 128 GPUs for
the largest dataset. Frontier is also a Slingshot 11 supercomputer with 9,856 compute nodes.  Each node
on Frontier has four AMD Instinct MI250X GPUs, each with 128 GB of HBM2E memory.
Each MI250X GPU is partitioned into two Graphic Compute Dies (GCDs) and each
GCD appears as a separate device for launching GPU kernels. The A100 GPU has a
peak of 19.5 FP32 Tflop/s, and the MI250X GPU has a peak of 47.9 FP32 Tflop/s.
There are four NICs per node on both systems with an injection bandwidth of 25 GB/s. We
use PyTorch Geometric 2.6.1, and PyTorch 2.6.0 with CUDA 12.4 on Perlmutter, and
ROCm 6.2.4 on Frontier.

\subsection{Description of Graph Datasets and the GNN}

We conduct experiments using graph datasets of varying sizes, as shown in
Table~\ref{tab:datasets}.  The Reddit dataset is available through PyTorch
Geometric, and contains post data from September 2014, with individual posts as
nodes and edges connecting two posts if the same user commented on
both~\cite{graphsage}.  The ogbn-products dataset is part of the Open Graph
Benchmark (OGB)~\cite{ogb}, and depicts Amazon's product co-purchasing network,
where nodes are products sold and edges indicate that the products are
purchased together.  The ogbn-papers100M dataset, also part of OGB, represents
the Microsoft Academic Graph (MAG), where nodes are papers and edges indicate
citation relationships.  For the Reddit, ogbn-products, and ogn-papers-100M
datasets, we used the input features and labels that were provided with the
datasets. 

The products-14M datasets is a larger Amazon products
network~\cite{ni-etal-2019-justifying}.  The Isolate-3-8M dataset is a subgraph
of a protein similarity network in HipMCL's data repository~\cite{hipmcl}. The
europe\_osm dataset, part of the 10th DIMACS Implementation
Challenge~\cite{europe_osm_dimacs10}, represents OpenStreetMap data for Europe,
where nodes correspond to geographical locations, and edges represent roads
connecting these points.  For the Isolate-3-8M, products-14M, and europe\_osm
datasets, we randomly generated input features with a size of 128, and
generated labels with 32 classes based on the distribution of node degrees.

\begin{table}[h]
    \caption{Details of graph datasets used for experiments.}
    \label{tab:datasets}
    \centering
    \resizebox{\columnwidth}{!}{
    \begin{tabular}{lrrrrrr}
        \toprule
        Dataset & \# Nodes & \# Edges & \# Non-zeros & \# Features & \# Classes \\
        \midrule
        Reddit        & 232,965     & 57,307,946 & 114,848,857 & 602  & 41  \\
        ogbn-products & 2,449,029   & 61,859,140 & 126,167,053  & 100  & 47  \\
        Isolate-3-8M  & 8,745,542   & 654,620,251 & 1,317,986,044  & 128  & 32  \\
        products-14M  & 14,249,639  & 115,394,635 & 245,036,907  & 128  & 32  \\
        europe\_osm   & 50,912,018  & 54,054,660 & 159,021,338  & 128  & 32  \\
        ogbn-papers100M & 111,059,956 & 1,615,685,872 & 1,726,745,828 & 100  & 172 \\
        \bottomrule
    \end{tabular}
    }
\end{table}

\begin{figure}[h]
  \centering
  \includegraphics[width=0.9\columnwidth]{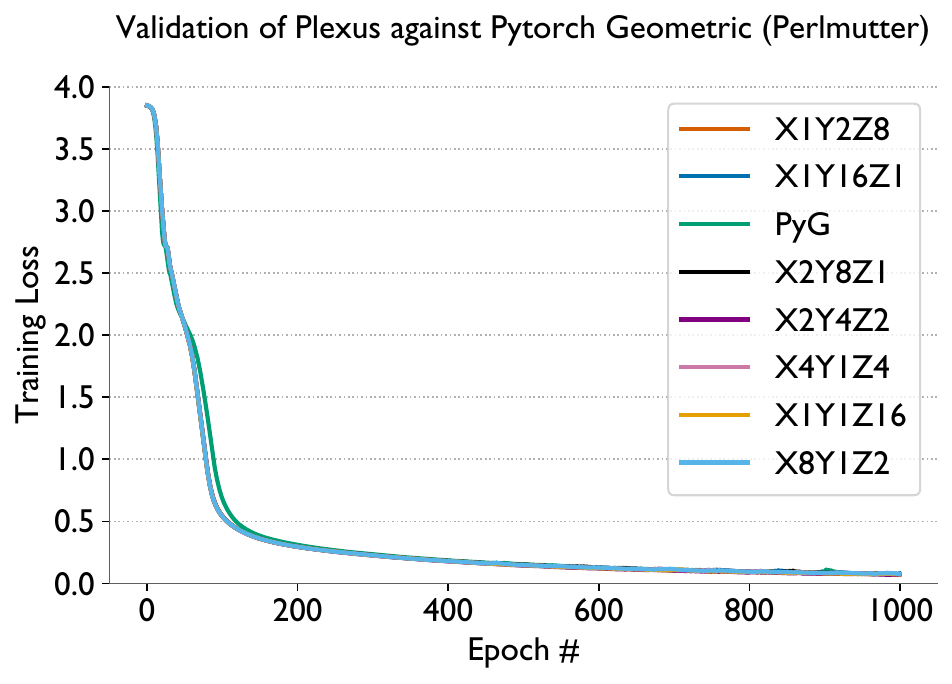}
  \caption{Validating \plexus against a serial PyTorch Geometric baseline on 16 GPUs of Perlmutter with ogbn-products.} \label{fig:val}
\end{figure}

For all the experiments, we create a GNN with three GCN layers and a hidden
dimension of 128, as increasing the model size beyond that has diminishing
returns on the model's generalization capabilities as shown in Jia et
al.~\cite{MLSYS2020ROC}. We train for ten epochs in each trial, and take the
average performance of the last eight epochs to account for initial
fluctuations. For each experiment, we run three independent trials and report
the average epoch time over three trials. We validated \plexus against PyTorch
Geometric for correctness as shown in Figure~\ref{fig:val}.

\subsection{Comparison with Other Frameworks}

We compare the performance of Plexus with that of SA, a sparsity-aware
implementation of CAGNET~\cite{cagnet-sparsity-aware}, and
BNS-GCN~\cite{bns-gcn}, two SOTA frameworks for distributed full-graph GNN
training that have previously been run on hundreds of GPUs as seen in
Table~\ref{tab:related-work}. We also compare with a variant of SA that uses
datasets partitioned using GVB~\cite{gvb-graph-partitioner}, a graph
partitioner used by the authors to improve performance (SA+GVB). We contacted
the authors to confirm that SA is the most recent and best performing
implementation of CAGNET.

For BNS-GCN, we use a boundary sampling rate of 1.0 since \plexus makes no
approximations and we are interested in comparing with similar settings. This
is akin to vanilla partition parallelism with METIS. We made a small
modification to the BNS-GCN code that resolved a bug that led to crashes during
training when the boundary size was 0. We also contacted the authors of BNS-GCN
regarding unexpectedly high runtimes with METIS, but did not receive a response
in time to compare against it.  We only compare with these frameworks on
Perlmutter as we encountered frequent stability issues and memory errors on
Frontier, preventing us from running experiments reliably.

\begin{figure*}[t]
  \centering
  \includegraphics[width=0.33\textwidth]{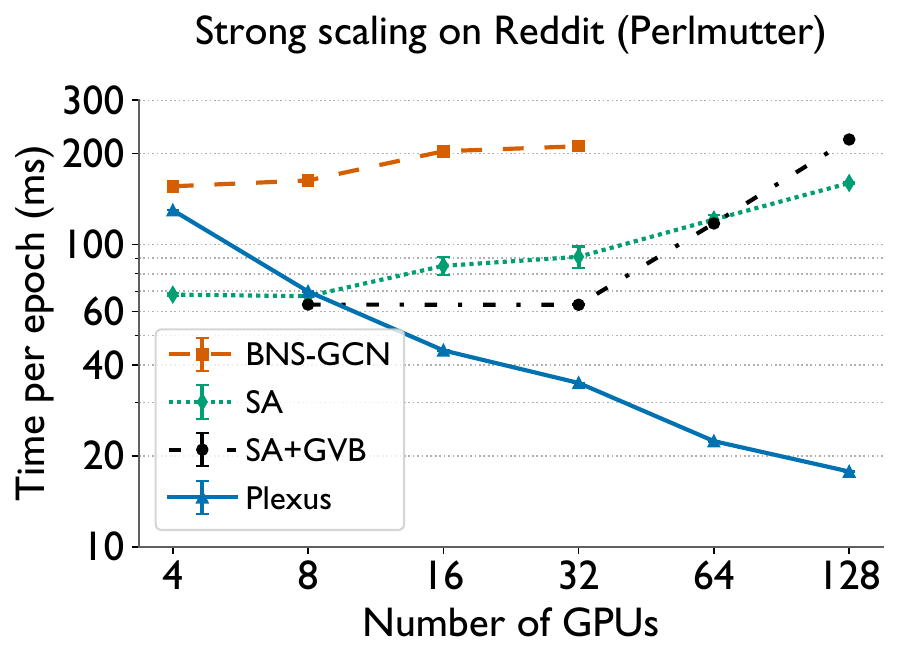}
  \includegraphics[width=0.33\textwidth]{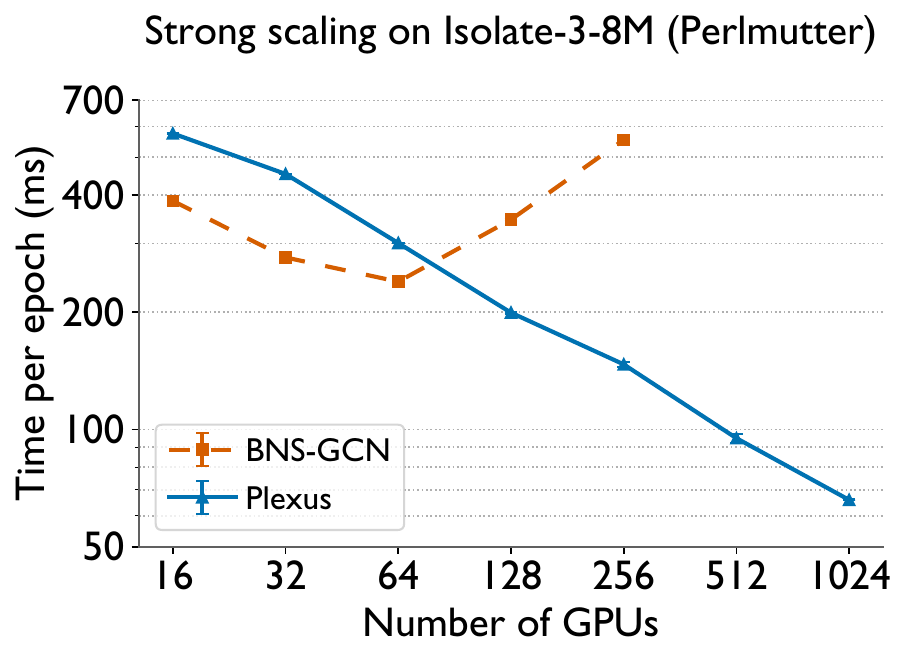}
  \includegraphics[width=0.33\textwidth]{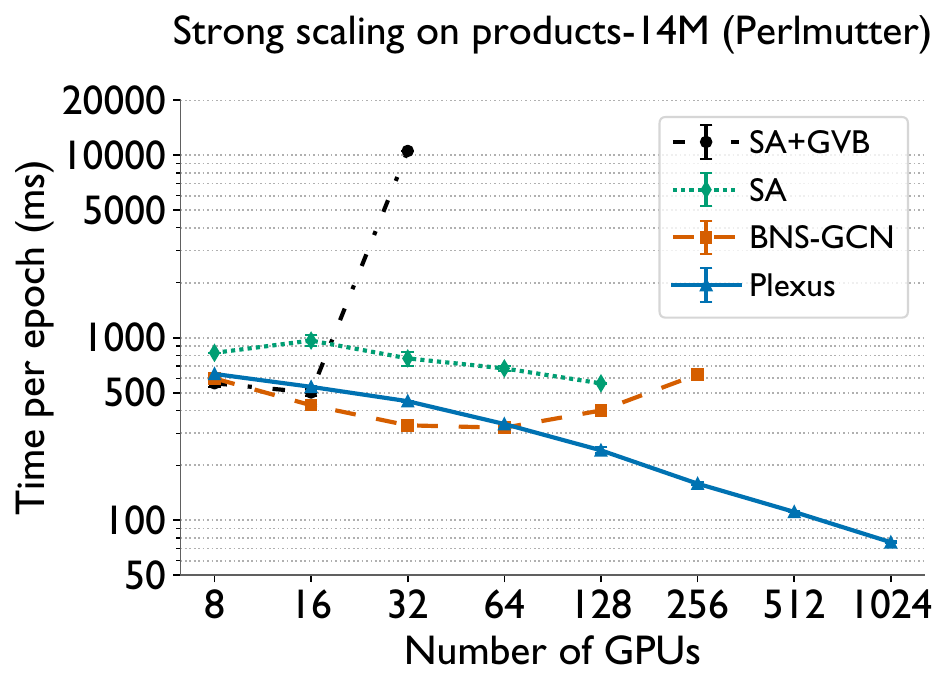}
  \caption{Comparison of strong scaling performance of \plexus, SA, SA+GVB, and BNS-GCN for several datasets on Perlmutter.}
  \label{fig:scaling-comp}
\end{figure*}

\section{Scaling Results}
\label{sec:results}
Finally, we present the results of our scaling experiments across six graph
datasets on both Perlmutter and Frontier, and compare \plexus with SA, SA+GVB,
and BNS-GCN.

\subsection{Comparison with SOTA Frameworks}

We compare \plexus to the other frameworks only on the Reddit, Isolate-3-8M,
and products-14M datasets. ogbn-papers100M results were limited due to
partitioning timeouts after 5 hours (BNS-GCN) and out-of-memory issues (SA,
SA+GVB). Figure~\ref{fig:scaling-comp} shows these comparative evaluation
results. For Reddit, SA performs better at 4 GPUs, but does not scale beyond
that. SA+GVB demonstrates somewhat better performance than SA upto 64 GPUs, but
also with poor scaling.  BNS-GCN scales similarly to SA but is slower in terms
of absolute time. \plexus is the only framework that achieves good strong
scaling up to 128 GPUs, and a 6$\times$ speedup over BNS-GCN on 32 GPUs and
9$\times$ over SA on 128 GPUs.

On Isolate-3-8M, both SA and SA+GVB failed to run due to out-of-memory issues.
BNS-GCN scales well to 64 GPUs, but quickly degrades in performance beyond this
point.  \plexus achieves a 3.8$\times$ speedup over BNS-GCN at 256 GPUs, and
scaling further to 1024 GPUs.  BNS-GCN's fine-grained communication is good at
a small scale, but has two key issues at larger scales. First, the partitioner
starts to divide denser subgraphs, resulting in a larger number of boundary
nodes. Second, BNS-GCN utilizes the all-to-all collective during communication.
Compared to ring-based collectives used in \plexus where GPUs only communicate
with their neighbors, all-to-alls send more long-distance messages, which leads
to higher latency. Without sampling boundary nodes, METIS is insufficient for
BNS-GCN to achieve comparable performance at scale.

For the products-14M dataset, we observe a similar pattern to Isolate-3-8M,
where BNS-GCN scales well till 64 GPUs, but then the performance drops sharply
following that. SA, on the other hand, starts off with a higher absolute time
but is able to scale comparatively better up to 128 GPUs. We tried running it
on 256 GPUs, but the job timed out at 20 minutes.  SA+GVB performs better than
SA for 8 and 16 GPUs, but has a drastic increase in time after that.  We
observe that \plexus scales up to 1024 GPUs and performs better than both
frameworks. It achieves a 2.3x speedup over SA on 128 GPUs and a 4x speedup
over BNS-GCN on 256 GPUs.

\begin{figure}[t]
  \centering
  \includegraphics[width=0.99\columnwidth]{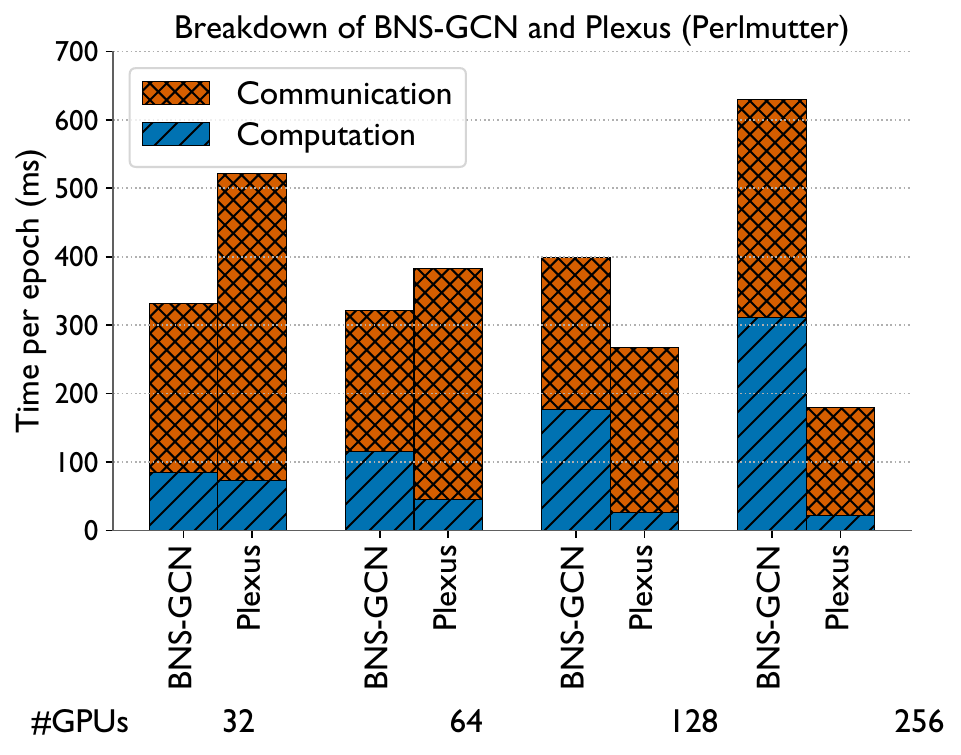}
  \caption{Breakdown of epoch times for BNS-GCN and \plexus on 32-256 GPUs of Perlmutter
  with products-14M.} \label{fig:breakdown}
\end{figure}

In order to understand the inflection point between BNS-GCN and \plexus at 64
GPUs further, we look at the breakdown of epoch times in
Figure~\ref{fig:breakdown}. At 32 GPUs, BNS-GCN completes an epoch faster than
\plexus primarily due to having a lower communication time, which can be
attributed to the fine-grained communication pattern of partition parallelism.
In \plexus, on the other hand, the communication time is higher since the
collectives are performed on the full dense outputs, and \plexus does not have
sparsity-aware modifications like SA. The inefficiency of the all-to-all
communication pattern employed by BNS-GCN becomes evident at 64 GPUs.

\begin{figure*}[t]
  \centering
  \includegraphics[width=\textwidth]{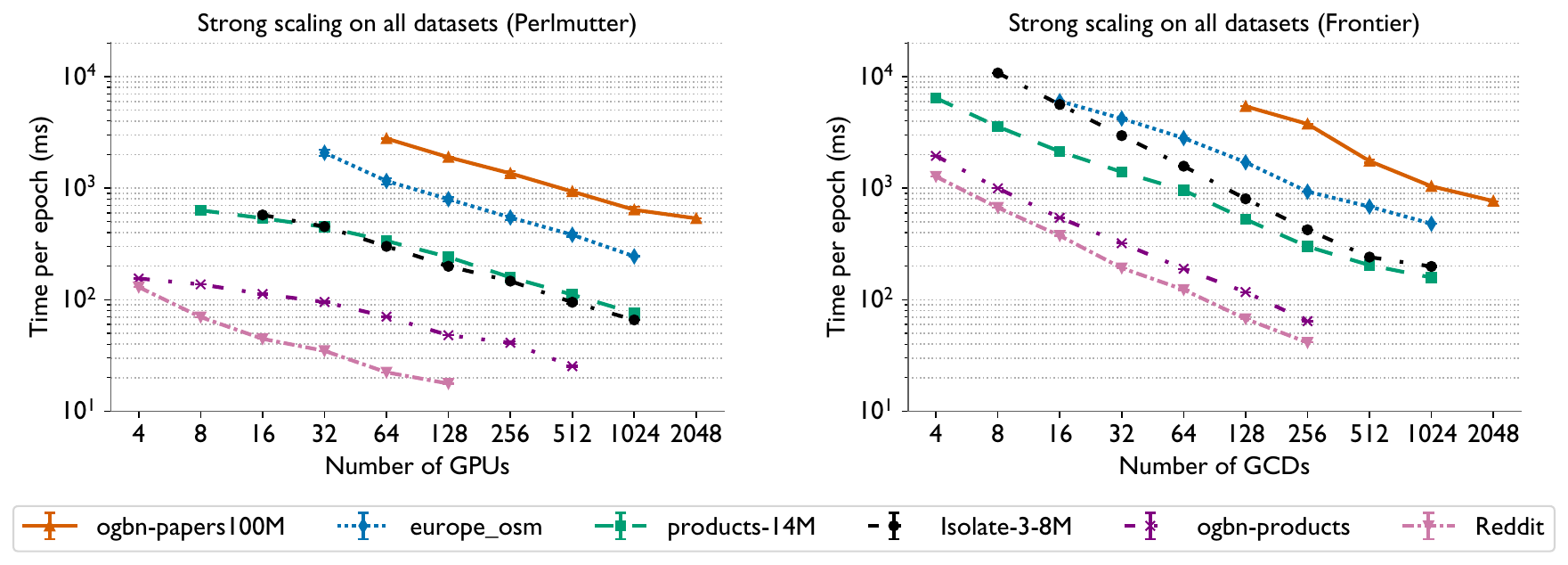}
  \caption{Strong scaling performance of \plexus for six graph datasets of
different sizes (Table~\ref{tab:datasets}) on both Perlmutter (left) and
Frontier (right). Note that the x-axis shows GPUs for Perlmutter and GCDs for
Frontier.}
  \label{fig:scaling-plexus}
\end{figure*}

Another interesting observation is that the computation scaling for the two
frameworks also differs. While \plexus shows notable improvements in the
computation time from 32 GPUs to 256 GPUs, BNS-GCN's computation time increases
with the number of GPUs. After further investigation, we found that the total
number of nodes across partitions, including boundary nodes, increased from 18M
to 22M for BNS-GCN when going from 32 to 256 GPUs. This explains why the local
computation of a partition also increases in addition to the poor scaling of
communication.

Overall, \plexus outperforms BNS-GCN, SA, and SA+GVB across the three datasets.
While BNS-GCN and SA are more efficient at small scales due to sparsity-aware
communication, they struggle at larger scales. \plexus scales well to 1024 GPUs
with the lowest absolute epoch times. Its scaling is also more consistent
across datasets, even performing competitively at small scales. All of this is
achieved without a graph partitioner. Unlike METIS, which timed out for some
datasets, and GVB, which ran out of memory on obgn-papers100M at 32 GPUs (as
noted by SA's authors in ~\cite{cagnet-sparsity-aware}), \plexus' double
permutation scheme mitigates load imbalance scalably with minimal overheads.

\subsection{Strong Scaling of \plexus}

In addition to the three datasets discussed above, we also ran \plexus on three
other datasets to demonstrate its strong scaling capabilities on both
Perlmutter and Frontier (Figure~\ref{fig:scaling-plexus}).  The sparsity level
of a graph determines the communication to computation ratio in \plexus.
As a result, \plexus scales better with Reddit, a denser graph compared to ogbn-products on Perlmutter (left plot).
When training with ogbn-products, \plexus becomes communication-dominated quicker than
Reddit, explaining the increasing gap between the performance for the two datasets (on Perlmutter).
This effect can similarly be seen with Isolate-3-8M and
products-14M. Even though products-14M has more nodes than Isolate-3-8M, the
latter is denser. This explains why \plexus is slower with Isolate-3-8M
at 16 GPUs where the computation cost is significant, but for products-14M, which is
more communication dominated, eventually \plexus takes longer beyond 64 GPUs. We also
show results for europe\_osm on 1024 GPUs and ogbn-papers100M on 2048 GPUs of
Perlmutter. We observe that the scaling with ogbn-papers100M starts to slow down at
2048 GPUs, at which point the computation cost is marginal. This is, to the
best of our knowledge, the largest number of GPUs that have been used for
parallel full-graph GNN training to date.

On Frontier (right plot), we notice generally better trends with all datasets
when compared to those on Perlmutter.  This is because the SpMM times on AMD GPUs were
an order of magnitude higher than on NVIDIA GPUs, allowing \plexus to scale
better. The trends observed on Perlmutter for Reddit and ogbn-products do not
hold here, and we do not observe a growing gap between the two datasets. Similarly,
\plexus is consistently slower with Isolate-3-8M than with products-14M since the former has a
higher number of edges. We also observe that \plexus demonstrates poorer scaling with europe\_osm, a sparser graph than both
products-14M and Isolate-3-8M.
Finally, we observe that \plexus demonstrates
impressive scaling for ogbn-papers100M, which is the largest graph dataset we ran with,
on up to 2048 GCDs.

\section{Conclusion}
\label{sec:conc}
GNN training has often relied on approximations such as mini-batch sampling due
to the high memory requirements of large graphs.  In the absence of efficient
and scalable full-graph alternatives, this approach has become the default in
many modern frameworks. In this work, we present \plexus, a three-dimensional
parallel framework for full-graph GNN training that adapts Agarwal et al.'s 3D
parallel matrix multiplication algorithm~\cite{agarwal-3d} to scale training
with billion-edge graphs to thousands of GPUs. \plexus includes a performance
model that selects an optimal 3D configuration based on communication and
computation costs, and incorporates several optimizations to further enhance
performance.  These include a double permutation scheme to reduce load
imbalance, and blocked aggregation to minimize variability. \plexus also offers
an easy-to-use API, eliminating the need for a graph partitioner and featuring
a parallel data loading utility that reduces CPU memory usage. Overall, this
work marks a significant step forward in making full-graph GNN training, a
notoriously challenging problem to scale, both practical and efficient.

\begin{acks}
This material is based upon work supported in part by the National Science
Foundation (NSF) under Grant No.~2047120. This research used resources of the
Oak Ridge Leadership Computing Facility at the Oak Ridge National Laboratory,
which is supported by the Office of Science of the U.S.~Department of Energy
under Contract No.~DE-AC05-00OR22725.  This research also used resources of the
National Energy Research Scientific Computing Center, a DOE Office of Science
User Facility supported by the Office of Science of the U.S.~Department of
Energy under Contract No.~DE-AC02-05CH11231 using NERSC award NERSC
DDR-ERCAP0034262.

\end{acks}

\bibliographystyle{ACM-Reference-Format.bst}
\bibliography{./bib/cite,./bib/pssg}


\begin{thebibliography}{52}


\ifx \showCODEN    \undefined \def \showCODEN     #1{\unskip}     \fi
\ifx \showISBNx    \undefined \def \showISBNx     #1{\unskip}     \fi
\ifx \showISBNxiii \undefined \def \showISBNxiii  #1{\unskip}     \fi
\ifx \showISSN     \undefined \def \showISSN      #1{\unskip}     \fi
\ifx \showLCCN     \undefined \def \showLCCN      #1{\unskip}     \fi
\ifx \shownote     \undefined \def \shownote      #1{#1}          \fi
\ifx \showarticletitle \undefined \def \showarticletitle #1{#1}   \fi
\ifx \showURL      \undefined \def \showURL       {\relax}        \fi
\providecommand\bibfield[2]{#2}
\providecommand\bibinfo[2]{#2}
\providecommand\natexlab[1]{#1}
\providecommand\showeprint[2][]{arXiv:#2}

\bibitem[osl(2021)]%
        {oslo}
 \bibinfo{year}{2021}\natexlab{}.
\newblock \bibinfo{title}{OSLO: Open Source for Large-scale Optimization}.
\newblock \bibinfo{howpublished}{\url{https://github.com/EleutherAI/oslo}}.
\newblock


\bibitem[Acer et~al\mbox{.}(2016)]%
        {gvb-graph-partitioner}
\bibfield{author}{\bibinfo{person}{Seher Acer}, \bibinfo{person}{Oguz
  Selvitopi}, {and} \bibinfo{person}{Cevdet Aykanat}.}
  \bibinfo{year}{2016}\natexlab{}.
\newblock \showarticletitle{Improving performance of sparse matrix dense matrix
  multiplication on large-scale parallel systems}.
\newblock \bibinfo{journal}{\emph{Parallel Comput.}} \bibinfo{volume}{59},
  \bibinfo{number}{C} (\bibinfo{date}{Nov.} \bibinfo{year}{2016}),
  \bibinfo{pages}{71–96}.
\newblock
\showISSN{0167-8191}
\href{https://doi.org/10.1016/j.parco.2016.10.001}{doi:\nolinkurl{10.1016/j.parco.2016.10.001}}


\bibitem[Agarwal et~al\mbox{.}(1995)]%
        {agarwal-3d}
\bibfield{author}{\bibinfo{person}{R.~C. Agarwal}, \bibinfo{person}{S.~M.
  Balle}, \bibinfo{person}{F.~G. Gustavson}, \bibinfo{person}{M. Joshi}, {and}
  \bibinfo{person}{P. Palkar}.} \bibinfo{year}{1995}\natexlab{}.
\newblock \showarticletitle{A three-dimensional approach to parallel matrix
  multiplication}.
\newblock \bibinfo{journal}{\emph{IBM Journal of Research and Development}}
  \bibinfo{volume}{39}, \bibinfo{number}{5} (\bibinfo{year}{1995}),
  \bibinfo{pages}{575--582}.
\newblock
\href{https://doi.org/10.1147/rd.395.0575}{doi:\nolinkurl{10.1147/rd.395.0575}}


\bibitem[Ai et~al\mbox{.}(2024)]%
        {neutrontp}
\bibfield{author}{\bibinfo{person}{Xin Ai}, \bibinfo{person}{Hao Yuan},
  \bibinfo{person}{Zeyu Ling}, \bibinfo{person}{Qiange Wang},
  \bibinfo{person}{Yanfeng Zhang}, \bibinfo{person}{Zhenbo Fu},
  \bibinfo{person}{Chaoyi Chen}, \bibinfo{person}{Yu Gu}, {and}
  \bibinfo{person}{Ge Yu}.} \bibinfo{year}{2024}\natexlab{}.
\newblock \bibinfo{title}{NeutronTP: Load-Balanced Distributed Full-Graph GNN
  Training with Tensor Parallelism}.
\newblock
\showeprint[arxiv]{2412.20379}~[cs.DC]
\urldef\tempurl%
\url{https://arxiv.org/abs/2412.20379}
\showURL{%
\tempurl}


\bibitem[Azad et~al\mbox{.}(2018)]%
        {hipmcl}
\bibfield{author}{\bibinfo{person}{Ariful Azad}, \bibinfo{person}{Georgios~A
  Pavlopoulos}, \bibinfo{person}{Christos~A Ouzounis}, \bibinfo{person}{Nikos~C
  Kyrpides}, {and} \bibinfo{person}{Aydin Buluç}.}
  \bibinfo{year}{2018}\natexlab{}.
\newblock \showarticletitle{HipMCL: a high-performance parallel implementation
  of the Markov clustering algorithm for large-scale networks}.
\newblock \bibinfo{journal}{\emph{Nucleic Acids Research}}
  \bibinfo{volume}{46}, \bibinfo{number}{6} (\bibinfo{date}{01}
  \bibinfo{year}{2018}), \bibinfo{pages}{e33--e33}.
\newblock
\showISSN{0305-1048}
\showeprint{https://academic.oup.com/nar/article-pdf/46/6/e33/24525991/gkx1313.pdf}
\href{https://doi.org/10.1093/nar/gkx1313}{doi:\nolinkurl{10.1093/nar/gkx1313}}


\bibitem[Balın et~al\mbox{.}(2021)]%
        {mg-gcn}
\bibfield{author}{\bibinfo{person}{Muhammed~Fatih Balın},
  \bibinfo{person}{Kaan Sancak}, {and} \bibinfo{person}{Ümit
  V.~Çatalyürek}.} \bibinfo{year}{2021}\natexlab{}.
\newblock \bibinfo{title}{MG-GCN: Scalable Multi-GPU GCN Training Framework}.
\newblock
\showeprint[arxiv]{2110.08688}~[cs.LG]
\urldef\tempurl%
\url{https://arxiv.org/abs/2110.08688}
\showURL{%
\tempurl}


\bibitem[Cai et~al\mbox{.}(2021)]%
        {dgcl}
\bibfield{author}{\bibinfo{person}{Zhenkun Cai}, \bibinfo{person}{Xiao Yan},
  \bibinfo{person}{Yidi Wu}, \bibinfo{person}{Kaihao Ma},
  \bibinfo{person}{James Cheng}, {and} \bibinfo{person}{Fan Yu}.}
  \bibinfo{year}{2021}\natexlab{}.
\newblock \showarticletitle{DGCL: an efficient communication library for
  distributed GNN training}. In \bibinfo{booktitle}{\emph{Proceedings of the
  Sixteenth European Conference on Computer Systems}} (Online Event, United
  Kingdom) \emph{(\bibinfo{series}{EuroSys '21})}.
  \bibinfo{publisher}{Association for Computing Machinery},
  \bibinfo{address}{New York, NY, USA}, \bibinfo{pages}{130–144}.
\newblock
\showISBNx{9781450383349}
\href{https://doi.org/10.1145/3447786.3456233}{doi:\nolinkurl{10.1145/3447786.3456233}}


\bibitem[Chen et~al\mbox{.}(2018a)]%
        {fastgcn}
\bibfield{author}{\bibinfo{person}{Jie Chen}, \bibinfo{person}{Tengfei Ma},
  {and} \bibinfo{person}{Cao Xiao}.} \bibinfo{year}{2018}\natexlab{a}.
\newblock \bibinfo{title}{FastGCN: Fast Learning with Graph Convolutional
  Networks via Importance Sampling}.
\newblock
\showeprint[arxiv]{1801.10247}~[cs.LG]
\urldef\tempurl%
\url{https://arxiv.org/abs/1801.10247}
\showURL{%
\tempurl}


\bibitem[Chen et~al\mbox{.}(2018b)]%
        {chen2018stochastictraininggraphconvolutional}
\bibfield{author}{\bibinfo{person}{Jianfei Chen}, \bibinfo{person}{Jun Zhu},
  {and} \bibinfo{person}{Le Song}.} \bibinfo{year}{2018}\natexlab{b}.
\newblock \bibinfo{title}{Stochastic Training of Graph Convolutional Networks
  with Variance Reduction}.
\newblock
\showeprint[arxiv]{1710.10568}~[stat.ML]
\urldef\tempurl%
\url{https://arxiv.org/abs/1710.10568}
\showURL{%
\tempurl}


\bibitem[Cheng et~al\mbox{.}(2023)]%
        {cheng2023atp}
\bibfield{author}{\bibinfo{person}{Shenggan Cheng}, \bibinfo{person}{Ziming
  Liu}, \bibinfo{person}{Jiangsu Du}, {and} \bibinfo{person}{Yang You}.}
  \bibinfo{year}{2023}\natexlab{}.
\newblock \showarticletitle{ATP: Adaptive Tensor Parallelism for Foundation
  Models}.
\newblock \bibinfo{journal}{\emph{arXiv preprint arXiv:2301.08658}}
  (\bibinfo{year}{2023}).
\newblock


\bibitem[Chiang et~al\mbox{.}(2019)]%
        {cluster-gcn}
\bibfield{author}{\bibinfo{person}{Wei-Lin Chiang}, \bibinfo{person}{Xuanqing
  Liu}, \bibinfo{person}{Si Si}, \bibinfo{person}{Yang Li},
  \bibinfo{person}{Samy Bengio}, {and} \bibinfo{person}{Cho-Jui Hsieh}.}
  \bibinfo{year}{2019}\natexlab{}.
\newblock \showarticletitle{Cluster-GCN: An Efficient Algorithm for Training
  Deep and Large Graph Convolutional Networks}. In
  \bibinfo{booktitle}{\emph{Proceedings of the 25th ACM SIGKDD International
  Conference on Knowledge Discovery \& Data Mining}}
  \emph{(\bibinfo{series}{KDD '19})}. \bibinfo{publisher}{ACM}.
\newblock
\href{https://doi.org/10.1145/3292500.3330925}{doi:\nolinkurl{10.1145/3292500.3330925}}


\bibitem[Das et~al\mbox{.}(2024)]%
        {ags-gnn}
\bibfield{author}{\bibinfo{person}{Siddhartha~Shankar Das},
  \bibinfo{person}{S~M Ferdous}, \bibinfo{person}{Mahantesh~M Halappanavar},
  \bibinfo{person}{Edoardo Serra}, {and} \bibinfo{person}{Alex Pothen}.}
  \bibinfo{year}{2024}\natexlab{}.
\newblock \bibinfo{title}{AGS-GNN: Attribute-guided Sampling for Graph Neural
  Networks}.
\newblock
\showeprint[arxiv]{2405.15218}~[cs.LG]
\urldef\tempurl%
\url{https://arxiv.org/abs/2405.15218}
\showURL{%
\tempurl}


\bibitem[Fey and Lenssen(2019)]%
        {pytorchgeometric}
\bibfield{author}{\bibinfo{person}{Matthias Fey} {and}
  \bibinfo{person}{Jan~Eric Lenssen}.} \bibinfo{year}{2019}\natexlab{}.
\newblock \bibinfo{title}{Fast Graph Representation Learning with PyTorch
  Geometric}.
\newblock
\showeprint[arxiv]{1903.02428}~[cs.LG]
\urldef\tempurl%
\url{https://arxiv.org/abs/1903.02428}
\showURL{%
\tempurl}


\bibitem[GmbH(2010)]%
        {europe_osm_dimacs10}
\bibfield{author}{\bibinfo{person}{Geofabrik GmbH}.}
  \bibinfo{year}{2010}\natexlab{}.
\newblock \bibinfo{title}{DIMACS10/europe\_osm}.
\newblock \bibinfo{howpublished}{SuiteSparse Matrix Collection}.
\newblock
\urldef\tempurl%
\url{https://sparse.tamu.edu/DIMACS10/europe_osm}
\showURL{%
\tempurl}


\bibitem[Hamilton et~al\mbox{.}(2018)]%
        {graphsage}
\bibfield{author}{\bibinfo{person}{William~L. Hamilton}, \bibinfo{person}{Rex
  Ying}, {and} \bibinfo{person}{Jure Leskovec}.}
  \bibinfo{year}{2018}\natexlab{}.
\newblock \bibinfo{title}{Inductive Representation Learning on Large Graphs}.
\newblock
\showeprint[arxiv]{1706.02216}~[cs.SI]
\urldef\tempurl%
\url{https://arxiv.org/abs/1706.02216}
\showURL{%
\tempurl}


\bibitem[Hu et~al\mbox{.}(2021)]%
        {ogb}
\bibfield{author}{\bibinfo{person}{Weihua Hu}, \bibinfo{person}{Matthias Fey},
  \bibinfo{person}{Marinka Zitnik}, \bibinfo{person}{Yuxiao Dong},
  \bibinfo{person}{Hongyu Ren}, \bibinfo{person}{Bowen Liu},
  \bibinfo{person}{Michele Catasta}, {and} \bibinfo{person}{Jure Leskovec}.}
  \bibinfo{year}{2021}\natexlab{}.
\newblock \bibinfo{title}{Open Graph Benchmark: Datasets for Machine Learning
  on Graphs}.
\newblock
\showeprint[arxiv]{2005.00687}~[cs.LG]
\urldef\tempurl%
\url{https://arxiv.org/abs/2005.00687}
\showURL{%
\tempurl}


\bibitem[Jia et~al\mbox{.}(2020)]%
        {MLSYS2020ROC}
\bibfield{author}{\bibinfo{person}{Zhihao Jia}, \bibinfo{person}{Sina Lin},
  \bibinfo{person}{Mingyu Gao}, \bibinfo{person}{Matei Zaharia}, {and}
  \bibinfo{person}{Alex Aiken}.} \bibinfo{year}{2020}\natexlab{}.
\newblock \showarticletitle{Improving the Accuracy, Scalability, and
  Performance of Graph Neural Networks with Roc}. In
  \bibinfo{booktitle}{\emph{Proceedings of Machine Learning and Systems}},
  \bibfield{editor}{\bibinfo{person}{I.~Dhillon},
  \bibinfo{person}{D.~Papailiopoulos}, {and} \bibinfo{person}{V.~Sze}} (Eds.),
  Vol.~\bibinfo{volume}{2}. \bibinfo{pages}{187--198}.
\newblock
\urldef\tempurl%
\url{https://proceedings.mlsys.org/paper\_files/paper/2020/file/91fc23ceccb664ebb0cf4257e1ba9c51-Paper.pdf}
\showURL{%
\tempurl}


\bibitem[Karypis and Kumar(1999)]%
        {metis}
\bibfield{author}{\bibinfo{person}{George Karypis} {and} \bibinfo{person}{Vipin
  Kumar}.} \bibinfo{year}{1999}\natexlab{}.
\newblock \showarticletitle{Kumar, V.: A Fast and High Quality Multilevel
  Scheme for Partitioning Irregular Graphs. SIAM Journal on Scientific
  Computing 20(1), 359-392}.
\newblock \bibinfo{journal}{\emph{Siam Journal on Scientific Computing}}
  \bibinfo{volume}{20} (\bibinfo{date}{01} \bibinfo{year}{1999}).
\newblock
\href{https://doi.org/10.1137/S1064827595287997}{doi:\nolinkurl{10.1137/S1064827595287997}}


\bibitem[Kipf and Welling(2016)]%
        {KipfW16}
\bibfield{author}{\bibinfo{person}{Thomas~N. Kipf} {and} \bibinfo{person}{Max
  Welling}.} \bibinfo{year}{2016}\natexlab{}.
\newblock \showarticletitle{Semi-Supervised Classification with Graph
  Convolutional Networks}.
\newblock \bibinfo{journal}{\emph{CoRR}}  \bibinfo{volume}{abs/1609.02907}
  (\bibinfo{year}{2016}).
\newblock
\showeprint{1609.02907}
\urldef\tempurl%
\url{http://arxiv.org/abs/1609.02907}
\showURL{%
\tempurl}


\bibitem[Kurt et~al\mbox{.}(2023)]%
        {rdm}
\bibfield{author}{\bibinfo{person}{Süreyya~Emre Kurt},
  \bibinfo{person}{Jinghua Yan}, \bibinfo{person}{Aravind Sukumaran-Rajam},
  \bibinfo{person}{Prashant Pandey}, {and} \bibinfo{person}{P. Sadayappan}.}
  \bibinfo{year}{2023}\natexlab{}.
\newblock \showarticletitle{Communication Optimization for Distributed
  Execution of Graph Neural Networks}. In \bibinfo{booktitle}{\emph{2023 IEEE
  International Parallel and Distributed Processing Symposium (IPDPS)}}.
  \bibinfo{pages}{512--523}.
\newblock
\href{https://doi.org/10.1109/IPDPS54959.2023.00058}{doi:\nolinkurl{10.1109/IPDPS54959.2023.00058}}


\bibitem[LeCun et~al\mbox{.}(1990)]%
        {lenet}
\bibfield{author}{\bibinfo{person}{Yann LeCun}, \bibinfo{person}{Bernhard
  Boser}, \bibinfo{person}{John Denker}, \bibinfo{person}{Donnie Henderson},
  \bibinfo{person}{R. Howard}, \bibinfo{person}{Wayne Hubbard}, {and}
  \bibinfo{person}{Lawrence Jackel}.} \bibinfo{year}{1990}\natexlab{}.
\newblock \showarticletitle{Handwritten Digit Recognition with a
  Back-Propagation Network}. In \bibinfo{booktitle}{\emph{Advances in Neural
  Information Processing Systems}},
  \bibfield{editor}{\bibinfo{person}{D.~Touretzky}} (Ed.),
  Vol.~\bibinfo{volume}{2}. \bibinfo{publisher}{Morgan-Kaufmann},
  \bibinfo{pages}{396--404}.
\newblock
\urldef\tempurl%
\url{https://proceedings.neurips.cc/paper/1989/file/53c3bce66e43be4f209556518c2fcb54-Paper.pdf}
\showURL{%
\tempurl}


\bibitem[Li et~al\mbox{.}(2018)]%
        {deeper-insights-gnn}
\bibfield{author}{\bibinfo{person}{Qimai Li}, \bibinfo{person}{Zhichao Han},
  {and} \bibinfo{person}{Xiao-Ming Wu}.} \bibinfo{year}{2018}\natexlab{}.
\newblock \showarticletitle{Deeper insights into graph convolutional networks
  for semi-supervised learning}. In \bibinfo{booktitle}{\emph{Proceedings of
  the Thirty-Second AAAI Conference on Artificial Intelligence and Thirtieth
  Innovative Applications of Artificial Intelligence Conference and Eighth AAAI
  Symposium on Educational Advances in Artificial Intelligence}} (New Orleans,
  Louisiana, USA) \emph{(\bibinfo{series}{AAAI'18/IAAI'18/EAAI'18})}.
  \bibinfo{publisher}{AAAI Press}, Article \bibinfo{articleno}{433},
  \bibinfo{numpages}{8}~pages.
\newblock
\showISBNx{978-1-57735-800-8}


\bibitem[Li et~al\mbox{.}(2023a)]%
        {li2023automated-oases}
\bibfield{author}{\bibinfo{person}{Shengwei Li}, \bibinfo{person}{Zhiquan Lai},
  \bibinfo{person}{Yanqi Hao}, \bibinfo{person}{Weijie Liu},
  \bibinfo{person}{Keshi Ge}, \bibinfo{person}{Xiaoge Deng},
  \bibinfo{person}{Dongsheng Li}, {and} \bibinfo{person}{Kai Lu}.}
  \bibinfo{year}{2023}\natexlab{a}.
\newblock \showarticletitle{Automated Tensor Model Parallelism with Overlapped
  Communication for Efficient Foundation Model Training}.
\newblock \bibinfo{journal}{\emph{arXiv preprint arXiv:2305.16121}}
  (\bibinfo{year}{2023}).
\newblock


\bibitem[Li et~al\mbox{.}(2023b)]%
        {colossalai2023unified}
\bibfield{author}{\bibinfo{person}{Shenggui Li}, \bibinfo{person}{Hongxin Liu},
  \bibinfo{person}{Zhengda Bian}, \bibinfo{person}{Jiarui Fang},
  \bibinfo{person}{Haichen Huang}, \bibinfo{person}{Yuliang Liu},
  \bibinfo{person}{Boxiang Wang}, {and} \bibinfo{person}{Yang You}.}
  \bibinfo{year}{2023}\natexlab{b}.
\newblock \showarticletitle{{Colossal-AI:} A Unified Deep Learning System For
  Large-Scale Parallel Training}. In \bibinfo{booktitle}{\emph{Proceedings of
  the 52nd International Conference on Parallel Processing}} (, Salt Lake City,
  UT, USA,) \emph{(\bibinfo{series}{ICPP '23})}.
  \bibinfo{publisher}{Association for Computing Machinery},
  \bibinfo{address}{New York, NY, USA}, \bibinfo{pages}{766–775}.
\newblock
\showISBNx{9798400708435}
\href{https://doi.org/10.1145/3605573.3605613}{doi:\nolinkurl{10.1145/3605573.3605613}}


\bibitem[Liu et~al\mbox{.}(2021)]%
        {liu2021samplingmethodsefficienttraining}
\bibfield{author}{\bibinfo{person}{Xin Liu}, \bibinfo{person}{Mingyu Yan},
  \bibinfo{person}{Lei Deng}, \bibinfo{person}{Guoqi Li},
  \bibinfo{person}{Xiaochun Ye}, {and} \bibinfo{person}{Dongrui Fan}.}
  \bibinfo{year}{2021}\natexlab{}.
\newblock \bibinfo{title}{Sampling methods for efficient training of graph
  convolutional networks: A survey}.
\newblock
\showeprint[arxiv]{2103.05872}~[cs.LG]
\urldef\tempurl%
\url{https://arxiv.org/abs/2103.05872}
\showURL{%
\tempurl}


\bibitem[Mukhopadhyay et~al\mbox{.}(2024)]%
        {cagnet-sparsity-aware}
\bibfield{author}{\bibinfo{person}{Ujjaini Mukhopadhyay}, \bibinfo{person}{Alok
  Tripathy}, \bibinfo{person}{Oguz Selvitopi}, \bibinfo{person}{Katherine
  Yelick}, {and} \bibinfo{person}{Aydin Buluc}.}
  \bibinfo{year}{2024}\natexlab{}.
\newblock \showarticletitle{Sparsity-Aware Communication for Distributed Graph
  Neural Network Training}. In \bibinfo{booktitle}{\emph{Proceedings of the
  53rd International Conference on Parallel Processing}} (Gotland, Sweden)
  \emph{(\bibinfo{series}{ICPP '24})}. \bibinfo{publisher}{Association for
  Computing Machinery}, \bibinfo{address}{New York, NY, USA},
  \bibinfo{pages}{117–126}.
\newblock
\showISBNx{9798400717932}
\href{https://doi.org/10.1145/3673038.3673152}{doi:\nolinkurl{10.1145/3673038.3673152}}


\bibitem[Ni et~al\mbox{.}(2019)]%
        {ni-etal-2019-justifying}
\bibfield{author}{\bibinfo{person}{Jianmo Ni}, \bibinfo{person}{Jiacheng Li},
  {and} \bibinfo{person}{Julian McAuley}.} \bibinfo{year}{2019}\natexlab{}.
\newblock \showarticletitle{Justifying Recommendations using Distantly-Labeled
  Reviews and Fine-Grained Aspects}. In \bibinfo{booktitle}{\emph{Proceedings
  of the 2019 Conference on Empirical Methods in Natural Language Processing
  and the 9th International Joint Conference on Natural Language Processing
  (EMNLP-IJCNLP)}}, \bibfield{editor}{\bibinfo{person}{Kentaro Inui},
  \bibinfo{person}{Jing Jiang}, \bibinfo{person}{Vincent Ng}, {and}
  \bibinfo{person}{Xiaojun Wan}} (Eds.). \bibinfo{publisher}{Association for
  Computational Linguistics}, \bibinfo{address}{Hong Kong, China},
  \bibinfo{pages}{188--197}.
\newblock
\href{https://doi.org/10.18653/v1/D19-1018}{doi:\nolinkurl{10.18653/v1/D19-1018}}


\bibitem[NVIDIA({[n.\,d.]})]%
        {ncu}
\bibfield{author}{\bibinfo{person}{NVIDIA}.}
  \bibinfo{year}{[n.\,d.]}\natexlab{}.
\newblock \bibinfo{title}{NVIDIA Nsight Compute}.
\newblock
  \bibinfo{howpublished}{\url{https://developer.nvidia.com/nsight-compute}}.
\newblock


\bibitem[Pedregosa et~al\mbox{.}(2011)]%
        {scikit-learn}
\bibfield{author}{\bibinfo{person}{F. Pedregosa}, \bibinfo{person}{G.
  Varoquaux}, \bibinfo{person}{A. Gramfort}, \bibinfo{person}{V. Michel},
  \bibinfo{person}{B. Thirion}, \bibinfo{person}{O. Grisel},
  \bibinfo{person}{M. Blondel}, \bibinfo{person}{P. Prettenhofer},
  \bibinfo{person}{R. Weiss}, \bibinfo{person}{V. Dubourg}, \bibinfo{person}{J.
  Vanderplas}, \bibinfo{person}{A. Passos}, \bibinfo{person}{D. Cournapeau},
  \bibinfo{person}{M. Brucher}, \bibinfo{person}{M. Perrot}, {and}
  \bibinfo{person}{E. Duchesnay}.} \bibinfo{year}{2011}\natexlab{}.
\newblock \showarticletitle{Scikit-learn: Machine Learning in {P}ython}.
\newblock \bibinfo{journal}{\emph{Journal of Machine Learning Research}}
  \bibinfo{volume}{12} (\bibinfo{year}{2011}), \bibinfo{pages}{2825--2830}.
\newblock


\bibitem[Peng et~al\mbox{.}(2022)]%
        {sancus}
\bibfield{author}{\bibinfo{person}{Jingshu Peng}, \bibinfo{person}{Zhao Chen},
  \bibinfo{person}{Yingxia Shao}, \bibinfo{person}{Yanyan Shen},
  \bibinfo{person}{Lei Chen}, {and} \bibinfo{person}{Jiannong Cao}.}
  \bibinfo{year}{2022}\natexlab{}.
\newblock \showarticletitle{Sancus: staleness-aware communication-avoiding
  full-graph decentralized training in large-scale graph neural networks}.
\newblock \bibinfo{journal}{\emph{Proc. VLDB Endow.}} \bibinfo{volume}{15},
  \bibinfo{number}{9} (\bibinfo{date}{May} \bibinfo{year}{2022}),
  \bibinfo{pages}{1937–1950}.
\newblock
\showISSN{2150-8097}
\href{https://doi.org/10.14778/3538598.3538614}{doi:\nolinkurl{10.14778/3538598.3538614}}


\bibitem[Rabenseifner(2004)]%
        {rabenseifneroptimization2004}
\bibfield{author}{\bibinfo{person}{Rolf Rabenseifner}.}
  \bibinfo{year}{2004}\natexlab{}.
\newblock \showarticletitle{Optimization of Collective Reduction Operations}.
  In \bibinfo{booktitle}{\emph{Computational Science - ICCS 2004}},
  \bibfield{editor}{\bibinfo{person}{Marian Bubak}, \bibinfo{person}{Geert~Dick
  van Albada}, \bibinfo{person}{Peter M.~A. Sloot}, {and} \bibinfo{person}{Jack
  Dongarra}} (Eds.). \bibinfo{publisher}{Springer Berlin Heidelberg},
  \bibinfo{address}{Berlin, Heidelberg}, \bibinfo{pages}{1--9}.
\newblock
\showISBNx{978-3-540-24685-5}


\bibitem[Selvitopi et~al\mbox{.}(2021)]%
        {distributed-memory-spmm}
\bibfield{author}{\bibinfo{person}{Oguz Selvitopi}, \bibinfo{person}{Benjamin
  Brock}, \bibinfo{person}{Israt Nisa}, \bibinfo{person}{Alok Tripathy},
  \bibinfo{person}{Katherine Yelick}, {and} \bibinfo{person}{Ayd\i{}n
  Bulu\c{c}}.} \bibinfo{year}{2021}\natexlab{}.
\newblock \showarticletitle{Distributed-memory parallel algorithms for sparse
  times tall-skinny-dense matrix multiplication}. In
  \bibinfo{booktitle}{\emph{Proceedings of the 35th ACM International
  Conference on Supercomputing}} (Virtual Event, USA)
  \emph{(\bibinfo{series}{ICS '21})}. \bibinfo{publisher}{Association for
  Computing Machinery}, \bibinfo{address}{New York, NY, USA},
  \bibinfo{pages}{431–442}.
\newblock
\showISBNx{9781450383356}
\href{https://doi.org/10.1145/3447818.3461472}{doi:\nolinkurl{10.1145/3447818.3461472}}


\bibitem[Shi et~al\mbox{.}(2017)]%
        {shi2017tnvnn}
\bibfield{author}{\bibinfo{person}{Shaohuai Shi}, \bibinfo{person}{Pengfei Xu},
  {and} \bibinfo{person}{Xiaowen Chu}.} \bibinfo{year}{2017}\natexlab{}.
\newblock \showarticletitle{Supervised Learning Based Algorithm Selection for
  Deep Neural Networks}. In \bibinfo{booktitle}{\emph{2017 IEEE 23rd
  International Conference on Parallel and Distributed Systems (ICPADS)}}.
  \bibinfo{pages}{344--351}.
\newblock
\href{https://doi.org/10.1109/ICPADS.2017.00053}{doi:\nolinkurl{10.1109/ICPADS.2017.00053}}


\bibitem[Singh and Bhatele(2022)]%
        {singh:ipdps2022}
\bibfield{author}{\bibinfo{person}{Siddharth Singh} {and}
  \bibinfo{person}{Abhinav Bhatele}.} \bibinfo{year}{2022}\natexlab{}.
\newblock \showarticletitle{{AxoNN}: An asynchronous, message-driven parallel
  framework for extreme-scale deep learning}. In
  \bibinfo{booktitle}{\emph{Proceedings of the IEEE International Parallel \&
  Distributed Processing Symposium}} \emph{(\bibinfo{series}{IPDPS '22})}.
  \bibinfo{publisher}{IEEE Computer Society}.
\newblock


\bibitem[Singh et~al\mbox{.}(2024)]%
        {singh:sc2024}
\bibfield{author}{\bibinfo{person}{Siddharth Singh}, \bibinfo{person}{Prajwal
  Singhania}, \bibinfo{person}{Aditya Ranjan}, \bibinfo{person}{John
  Kirchenbauer}, \bibinfo{person}{Jonas Geiping}, \bibinfo{person}{Yuxin Wen},
  \bibinfo{person}{Neel Jain}, \bibinfo{person}{Abhimanyu Hans},
  \bibinfo{person}{Manli Shu}, \bibinfo{person}{Aditya Tomar},
  \bibinfo{person}{Tom Goldstein}, {and} \bibinfo{person}{Abhinav Bhatele}.}
  \bibinfo{year}{2024}\natexlab{}.
\newblock \showarticletitle{Democratizing {AI}: Open-source Scalable {LLM}
  Training on {GPU}-based Supercomputers}. In
  \bibinfo{booktitle}{\emph{Proceedings of the ACM/IEEE International
  Conference for High Performance Computing, Networking, Storage and Analysis}}
  \emph{(\bibinfo{series}{SC '24})}.
\newblock


\bibitem[Song et~al\mbox{.}(2024)]%
        {granndis}
\bibfield{author}{\bibinfo{person}{Jaeyong Song}, \bibinfo{person}{Hongsun
  Jang}, \bibinfo{person}{Jaewon Jung}, \bibinfo{person}{Youngsok Kim}, {and}
  \bibinfo{person}{Jinho Lee}.} \bibinfo{year}{2024}\natexlab{}.
\newblock \bibinfo{title}{GraNNDis: Efficient Unified Distributed Training
  Framework for Deep GNNs on Large Clusters}.
\newblock
\showeprint[arxiv]{2311.06837}~[cs.LG]
\urldef\tempurl%
\url{https://arxiv.org/abs/2311.06837}
\showURL{%
\tempurl}


\bibitem[Thakur and Gropp(2003)]%
        {thakurimproving2003}
\bibfield{author}{\bibinfo{person}{Rajeev Thakur} {and}
  \bibinfo{person}{William~D. Gropp}.} \bibinfo{year}{2003}\natexlab{}.
\newblock \showarticletitle{Improving the Performance of Collective Operations
  in MPICH}. In \bibinfo{booktitle}{\emph{Recent Advances in Parallel Virtual
  Machine and Message Passing Interface}},
  \bibfield{editor}{\bibinfo{person}{Jack Dongarra}, \bibinfo{person}{Domenico
  Laforenza}, {and} \bibinfo{person}{Salvatore Orlando}} (Eds.).
  \bibinfo{publisher}{Springer Berlin Heidelberg}, \bibinfo{address}{Berlin,
  Heidelberg}, \bibinfo{pages}{257--267}.
\newblock
\showISBNx{978-3-540-39924-7}


\bibitem[Tripathy et~al\mbox{.}(2020)]%
        {cagnet}
\bibfield{author}{\bibinfo{person}{Alok Tripathy}, \bibinfo{person}{Katherine
  Yelick}, {and} \bibinfo{person}{Ayd\i{}n Bulu\c{c}}.}
  \bibinfo{year}{2020}\natexlab{}.
\newblock \showarticletitle{Reducing communication in graph neural network
  training}. In \bibinfo{booktitle}{\emph{Proceedings of the International
  Conference for High Performance Computing, Networking, Storage and Analysis}}
  (Atlanta, Georgia) \emph{(\bibinfo{series}{SC '20})}.
  \bibinfo{publisher}{IEEE Press}, Article \bibinfo{articleno}{70},
  \bibinfo{numpages}{17}~pages.
\newblock
\showISBNx{9781728199986}


\bibitem[Veličković et~al\mbox{.}(2018)]%
        {GAT}
\bibfield{author}{\bibinfo{person}{Petar Veličković},
  \bibinfo{person}{Guillem Cucurull}, \bibinfo{person}{Arantxa Casanova},
  \bibinfo{person}{Adriana Romero}, \bibinfo{person}{Pietro Liò}, {and}
  \bibinfo{person}{Yoshua Bengio}.} \bibinfo{year}{2018}\natexlab{}.
\newblock \bibinfo{title}{Graph Attention Networks}.
\newblock
\showeprint[arxiv]{1710.10903}~[stat.ML]
\urldef\tempurl%
\url{https://arxiv.org/abs/1710.10903}
\showURL{%
\tempurl}


\bibitem[Wan et~al\mbox{.}(2023)]%
        {adaqp}
\bibfield{author}{\bibinfo{person}{Borui Wan}, \bibinfo{person}{Juntao Zhao},
  {and} \bibinfo{person}{Chuan Wu}.} \bibinfo{year}{2023}\natexlab{}.
\newblock \bibinfo{title}{Adaptive Message Quantization and Parallelization for
  Distributed Full-graph GNN Training}.
\newblock
\showeprint[arxiv]{2306.01381}~[cs.LG]
\urldef\tempurl%
\url{https://arxiv.org/abs/2306.01381}
\showURL{%
\tempurl}


\bibitem[Wan et~al\mbox{.}(2022a)]%
        {bns-gcn}
\bibfield{author}{\bibinfo{person}{Cheng Wan}, \bibinfo{person}{Youjie Li},
  \bibinfo{person}{Ang Li}, \bibinfo{person}{Nam~Sung Kim}, {and}
  \bibinfo{person}{Yingyan Lin}.} \bibinfo{year}{2022}\natexlab{a}.
\newblock \bibinfo{title}{BNS-GCN: Efficient Full-Graph Training of Graph
  Convolutional Networks with Partition-Parallelism and Random Boundary Node
  Sampling}.
\newblock
\showeprint[arxiv]{2203.10983}~[cs.LG]
\urldef\tempurl%
\url{https://arxiv.org/abs/2203.10983}
\showURL{%
\tempurl}


\bibitem[Wan et~al\mbox{.}(2022b)]%
        {pipegcn}
\bibfield{author}{\bibinfo{person}{Cheng Wan}, \bibinfo{person}{Youjie Li},
  \bibinfo{person}{Cameron~R. Wolfe}, \bibinfo{person}{Anastasios Kyrillidis},
  \bibinfo{person}{Nam~Sung Kim}, {and} \bibinfo{person}{Yingyan Lin}.}
  \bibinfo{year}{2022}\natexlab{b}.
\newblock \bibinfo{title}{PipeGCN: Efficient Full-Graph Training of Graph
  Convolutional Networks with Pipelined Feature Communication}.
\newblock
\showeprint[arxiv]{2203.10428}~[cs.LG]
\urldef\tempurl%
\url{https://arxiv.org/abs/2203.10428}
\showURL{%
\tempurl}


\bibitem[Wang et~al\mbox{.}(2020)]%
        {dgl}
\bibfield{author}{\bibinfo{person}{Minjie Wang}, \bibinfo{person}{Da Zheng},
  \bibinfo{person}{Zihao Ye}, \bibinfo{person}{Quan Gan},
  \bibinfo{person}{Mufei Li}, \bibinfo{person}{Xiang Song},
  \bibinfo{person}{Jinjing Zhou}, \bibinfo{person}{Chao Ma},
  \bibinfo{person}{Lingfan Yu}, \bibinfo{person}{Yu Gai},
  \bibinfo{person}{Tianjun Xiao}, \bibinfo{person}{Tong He},
  \bibinfo{person}{George Karypis}, \bibinfo{person}{Jinyang Li}, {and}
  \bibinfo{person}{Zheng Zhang}.} \bibinfo{year}{2020}\natexlab{}.
\newblock \bibinfo{title}{Deep Graph Library: A Graph-Centric,
  Highly-Performant Package for Graph Neural Networks}.
\newblock
\showeprint[arxiv]{1909.01315}~[cs.LG]
\urldef\tempurl%
\url{https://arxiv.org/abs/1909.01315}
\showURL{%
\tempurl}


\bibitem[Wang et~al\mbox{.}(2022)]%
        {neutronstar}
\bibfield{author}{\bibinfo{person}{Qiange Wang}, \bibinfo{person}{Yanfeng
  Zhang}, \bibinfo{person}{Hao Wang}, \bibinfo{person}{Chaoyi Chen},
  \bibinfo{person}{Xiaodong Zhang}, {and} \bibinfo{person}{Ge Yu}.}
  \bibinfo{year}{2022}\natexlab{}.
\newblock \showarticletitle{NeutronStar: Distributed GNN Training with Hybrid
  Dependency Management}. In \bibinfo{booktitle}{\emph{Proceedings of the 2022
  International Conference on Management of Data}} (Philadelphia, PA, USA)
  \emph{(\bibinfo{series}{SIGMOD '22})}. \bibinfo{publisher}{Association for
  Computing Machinery}, \bibinfo{address}{New York, NY, USA},
  \bibinfo{pages}{1301–1315}.
\newblock
\showISBNx{9781450392495}
\href{https://doi.org/10.1145/3514221.3526134}{doi:\nolinkurl{10.1145/3514221.3526134}}


\bibitem[Wang et~al\mbox{.}(2023)]%
        {mgg}
\bibfield{author}{\bibinfo{person}{Yuke Wang}, \bibinfo{person}{Boyuan Feng},
  \bibinfo{person}{Zheng Wang}, \bibinfo{person}{Tong Geng},
  \bibinfo{person}{Kevin Barker}, \bibinfo{person}{Ang Li}, {and}
  \bibinfo{person}{Yufei Ding}.} \bibinfo{year}{2023}\natexlab{}.
\newblock \showarticletitle{{MGG}: Accelerating Graph Neural Networks with
  {Fine-Grained} {Intra-Kernel} {Communication-Computation} Pipelining on
  {Multi-GPU} Platforms}. In \bibinfo{booktitle}{\emph{17th USENIX Symposium on
  Operating Systems Design and Implementation (OSDI 23)}}.
  \bibinfo{publisher}{USENIX Association}, \bibinfo{address}{Boston, MA},
  \bibinfo{pages}{779--795}.
\newblock
\showISBNx{978-1-939133-34-2}
\urldef\tempurl%
\url{https://www.usenix.org/conference/osdi23/presentation/wang-yuke}
\showURL{%
\tempurl}


\bibitem[Xu et~al\mbox{.}(2019)]%
        {GIN}
\bibfield{author}{\bibinfo{person}{Keyulu Xu}, \bibinfo{person}{Weihua Hu},
  \bibinfo{person}{Jure Leskovec}, {and} \bibinfo{person}{Stefanie Jegelka}.}
  \bibinfo{year}{2019}\natexlab{}.
\newblock \bibinfo{title}{How Powerful are Graph Neural Networks?}
\newblock
\showeprint[arxiv]{1810.00826}~[cs.LG]
\urldef\tempurl%
\url{https://arxiv.org/abs/1810.00826}
\showURL{%
\tempurl}


\bibitem[Yang et~al\mbox{.}(2018)]%
        {sparse-design-principles}
\bibfield{author}{\bibinfo{person}{Carl Yang}, \bibinfo{person}{Aydin Buluc},
  {and} \bibinfo{person}{John~D. Owens}.} \bibinfo{year}{2018}\natexlab{}.
\newblock \bibinfo{title}{Design Principles for Sparse Matrix Multiplication on
  the GPU}.
\newblock
\href{https://doi.org/10.48550/ARXIV.1803.08601}{doi:\nolinkurl{10.48550/ARXIV.1803.08601}}


\bibitem[Younesian et~al\mbox{.}(2024)]%
        {grapes}
\bibfield{author}{\bibinfo{person}{Taraneh Younesian}, \bibinfo{person}{Daniel
  Daza}, \bibinfo{person}{Emile van Krieken}, \bibinfo{person}{Thiviyan
  Thanapalasingam}, {and} \bibinfo{person}{Peter Bloem}.}
  \bibinfo{year}{2024}\natexlab{}.
\newblock \bibinfo{title}{GRAPES: Learning to Sample Graphs for Scalable Graph
  Neural Networks}.
\newblock
\showeprint[arxiv]{2310.03399}~[cs.LG]
\urldef\tempurl%
\url{https://arxiv.org/abs/2310.03399}
\showURL{%
\tempurl}


\bibitem[Yuan et~al\mbox{.}(2024)]%
        {yuan2024comprehensiveevaluationgnntraining}
\bibfield{author}{\bibinfo{person}{Hao Yuan}, \bibinfo{person}{Yajiong Liu},
  \bibinfo{person}{Yanfeng Zhang}, \bibinfo{person}{Xin Ai},
  \bibinfo{person}{Qiange Wang}, \bibinfo{person}{Chaoyi Chen},
  \bibinfo{person}{Yu Gu}, {and} \bibinfo{person}{Ge Yu}.}
  \bibinfo{year}{2024}\natexlab{}.
\newblock \bibinfo{title}{Comprehensive Evaluation of GNN Training Systems: A
  Data Management Perspective}.
\newblock
\showeprint[arxiv]{2311.13279}~[cs.LG]
\urldef\tempurl%
\url{https://arxiv.org/abs/2311.13279}
\showURL{%
\tempurl}


\bibitem[Zhang et~al\mbox{.}(2024)]%
        {cdfgnn}
\bibfield{author}{\bibinfo{person}{Shuai Zhang}, \bibinfo{person}{Zite Jiang},
  {and} \bibinfo{person}{Haihang You}.} \bibinfo{year}{2024}\natexlab{}.
\newblock \bibinfo{title}{CDFGNN: a Systematic Design of Cache-based
  Distributed Full-Batch Graph Neural Network Training with Communication
  Reduction}.
\newblock
\showeprint[arxiv]{2408.00232}~[cs.DC]
\urldef\tempurl%
\url{https://arxiv.org/abs/2408.00232}
\showURL{%
\tempurl}


\bibitem[Zheng et~al\mbox{.}(2022)]%
        {alpa}
\bibfield{author}{\bibinfo{person}{Lianmin Zheng}, \bibinfo{person}{Zhuohan
  Li}, \bibinfo{person}{Hao Zhang}, \bibinfo{person}{Yonghao Zhuang},
  \bibinfo{person}{Zhifeng Chen}, \bibinfo{person}{Yanping Huang},
  \bibinfo{person}{Yida Wang}, \bibinfo{person}{Yuanzhong Xu},
  \bibinfo{person}{Danyang Zhuo}, \bibinfo{person}{Joseph~E. Gonzalez}, {and}
  \bibinfo{person}{Ion Stoica}.} \bibinfo{year}{2022}\natexlab{}.
\newblock \showarticletitle{Alpa: Automating Inter- and Intra-Operator
  Parallelism for Distributed Deep Learning}.
\newblock \bibinfo{journal}{\emph{CoRR}}  \bibinfo{volume}{abs/2201.12023}
  (\bibinfo{year}{2022}).
\newblock
\showeprint[arXiv]{2201.12023}


\bibitem[Zou et~al\mbox{.}(2019)]%
        {ladies}
\bibfield{author}{\bibinfo{person}{Difan Zou}, \bibinfo{person}{Ziniu Hu},
  \bibinfo{person}{Yewen Wang}, \bibinfo{person}{Song Jiang},
  \bibinfo{person}{Yizhou Sun}, {and} \bibinfo{person}{Quanquan Gu}.}
  \bibinfo{year}{2019}\natexlab{}.
\newblock \bibinfo{title}{Layer-Dependent Importance Sampling for Training Deep
  and Large Graph Convolutional Networks}.
\newblock
\showeprint[arxiv]{1911.07323}~[cs.LG]
\urldef\tempurl%
\url{https://arxiv.org/abs/1911.07323}
\showURL{%
\tempurl}


\end{thebibliography}

\end{document}